\documentclass[10pt,twocolumn,letterpaper]{article}

\usepackage{authblk}

\usepackage[pagenumbers]{cvpr} 

\usepackage[dvipsnames]{xcolor}
\definecolor{cvprblue}{rgb}{0.21,0.49,0.74}
\usepackage[pagebackref,breaklinks,colorlinks,citecolor=cvprblue]{hyperref}
\usepackage{xcolor,enumitem,cite}

\AtBeginDocument{%
  \providecommand\BibTeX{{%
    \normalfont B\kern-0.5em{\scshape i\kern-0.25em b}\kern-0.8em\TeX}}}

\usepackage{booktabs} 
\usepackage{amsmath}

\DeclareMathOperator*{\argmin}{argmin}

\usepackage{pifont}

\newcommand{\cmark}{\textcolor{green}{\text{\ding{51}}}}%
\newcommand{\xmark}{\textcolor{red}{\text{\ding{55}}}}%

\def\paperID{?} 
\def\confName{}
\def\confYear{2024}

\author{\vspace{-0.4cm} Jia Wei Sii}
\author{Chee Seng Chan}

\affil{Center of Signal and Image Processing (CISiP), Universiti Malaya, 50603 Kuala Lumpur, Malaysia}
\affil[ ]{\texttt{\small \{17069558@siswa.um.edu.my,cs.chan@um.edu.my\}}} 
\affil[ ]{\small \url{https://github.com/JiaWeiSii/gorgeous/}}

\begin{document}

\title{Gorgeous: Create Your Desired Character Facial Makeup from Any Ideas}

\twocolumn[{%
\renewcommand\twocolumn[1][]{#1}%
\maketitle

\begin{center}
    \centering
    \vspace{-0.4cm}
    \includegraphics[width=\textwidth]{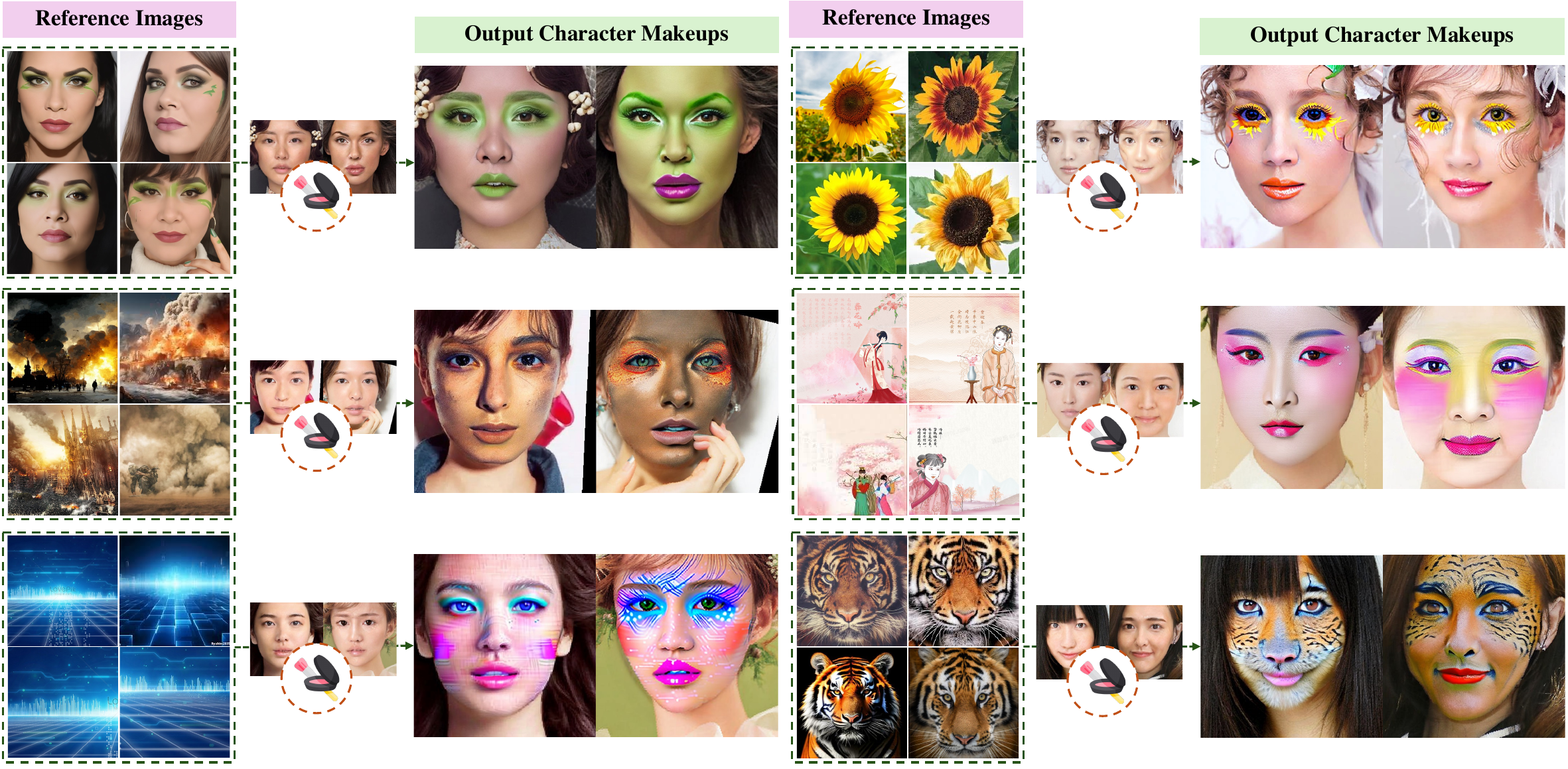}
    \captionof{figure}{Provide us with \textit{any} reference images of your desired character settings (e.g., war, sunflower), and our $Gorgeous$ will transform them into a creative and unique character makeup design that enriches your visual storytelling!
    }
    \label{fig:teaser}
\end{center}%
}]


\begin{abstract}

Contemporary makeup transfer methods primarily focus on replicating makeup from one face to another, considerably limiting their use in creating diverse and creative character makeup essential for visual storytelling. Such methods typically fail to address the need for uniqueness and contextual relevance, specifically aligning with character and story settings as they depend heavily on existing facial makeup in reference images. This approach also presents a significant challenge when attempting to source a perfectly matched facial makeup style, further complicating the creation of makeup designs inspired by various story elements, such as theme, background, and props that do not necessarily feature faces.
To address these limitations, we introduce $Gorgeous$, a novel diffusion-based makeup application method that goes beyond simple transfer by innovatively crafting unique and thematic facial makeup. Unlike traditional methods, $Gorgeous$ does not require the presence of a face in the reference images. Instead, it draws artistic inspiration from a minimal set of three to five images, which can be of any type, and transforms these elements into practical makeup applications directly on the face.
Our comprehensive experiments demonstrate that $Gorgeous$ can effectively generate distinctive character facial makeup inspired by the chosen thematic reference images. This approach opens up new possibilities for integrating broader story elements into character makeup, thereby enhancing the narrative depth and visual impact in storytelling.
%

\end{abstract}

\maketitle

\section{Introduction}

Character facial makeup can dramatically enhance visual storytelling in a movie, theatre productions, cosplay events, fashion shows, theme parks and more. For instance, facial makeup with gritty, metallic tones and rough textures strongly reflects the harsh environment in a war movie. These special effects makeup not only create visually stunning character transformations but also bring fantastical elements to life, making the imaginary worlds more tangible and believable to the audience. Typically, makeup artists carefully design these looks to complement every aspect of the story, ensuring that each character's appearance aligns with their personality and circumstances. However, applying these intricate makeup designs can be quite costly and time-consuming. Fortunately, makeup transfer technology provides tools to quickly and cost-effectively visualize these effects. This technology allows filmmakers and artists to experiment with various makeup looks digitally before committing to applying them physically, streamlining the creative process and reducing production costs.

Existing makeup transfer methods \cite{tong2007example,guo2009digital,liu2016makeup,li2018beautygan,chang2018pairedcyclegan,chen2019beautyglow,jiang2020psgan,deng2021spatially,nguyen2021lipstick,lyu2021sogan,yang2022elegant,sun2022ssat,yan2023beautyrec} primarily focus on ``replicating'' the makeup from a source face onto a target face, which inherently requires a source face to begin with. This limitation means that truly digital visualization is restricted, as there is still a need to ``create'' the data for transferring; it involves no inherent creativity, simply transferring existing designs. From our point of view, makeup design should be able to draw inspiration from any sources, not only makeup images but also natural elements such as animals or even broader themes such as a series of photographs depicting war (Fig.~\ref{fig:teaser}). Current makeup transfer methods fall short when the inspirations lack a direct facial representation. 

To bridge these gaps, we propose $Gorgeous$, a novel makeup application method to specially create unique and creative facial makeup for any themed characters, from a minimum set of 3 to 5 images. Unlike conventional makeup transfer methods, $Gorgeous$ does not merely copy makeup from one face to another. Instead, it allows for inspiration from any image. Also, the source images do not need to feature a face but can be any image type that embodies desired inspirational elements (see Fig.~\ref{fig:teaser}). 

Our solution, $Gorgeous$ is built on top of diffusion models \cite{Rombach_2022_CVPR,Zhang2023AddingCC,Gal2022AnII,Meng2021SDEditGI} and consists of three components: \textbf{(i) Makeup Formatting (MaFor) Module:} Utilizing ControlNet, trained on pseudo-paired makeup datasets, this module is equipped with essential makeup knowledge such as lipstick, eyeshadow, blushes, and foundation. Moreover, it is designed to maintain the individual's facial identity throughout the makeup application process. \textbf{(ii) Character Settings Learning (CSL) Module:} Leverages textual inversion to learn and encode artistic elements from a few inspirational reference images into textual embeddings for makeup styles. \textbf{(iii) Makeup Inpainting Pipeline (MaIP):} Adapts the idea of image inpainting to focus makeup application on facial areas while preserving the integrity of the non-facial regions through effective masking during the denoising process.

Our contributions are as follows: \textbf{(i) Novel Makeup Application:} We pioneer a method named $Gorgeous$ for creative and distinct character facial makeups from various inspiration sources. \textbf{(ii) Flexible Makeup Formatting:} Through the $MaFor$ module, $Gorgeous$ can transform any artistic concept into a practical makeup format, facilitating easy and versatile makeup creation. \textbf{(iii) Focused Makeup Inpainting:} The $MaIP$ module ensures makeup is applied precisely where needed, preserving the original aesthetics of the non-facial areas.  Comprehensive experiments validate $Gorgeous$'s effectiveness, demonstrating superior performance in creating new and engaging character makeups, surpassing conventional approaches in both qualitative and quantitative assessments.

\section{Related Works}
\subsection{Facial Makeup}
Existing methods such as traditional face-warping makeup methods \cite{guo2009digital,liu2016makeup} and GAN-based \cite{chang2018pairedcyclegan,li2018beautygan,chen2019beautyglow,deng2021spatially,yang2022elegant,jiang2020psgan,lyu2021sogan,nguyen2021lipstick,sun2022ssat,yan2023beautyrec} and diffusion-based \cite{Jin2024TowardTA} makeup transfer approaches primarily focus on duplicating makeup from one face to another. These methods are limited in their ability to create personalized and unique makeup designs, as they require an existing makeup look on a source face to initiate the transfer. In contrast, makeup recommendation systems \cite{nguyen2017smart,alashkar2017examples,alashkar2017rule,perera2021virtual,gulati2023beautifai} provide suggestions based on predefined rules and user characteristics. However, they do not generate new makeup styles from scratch but rather suggest variations based on existing templates and user input, which can still restrict the creativity necessary for fully personalized character design in visual storytelling. 

\subsection{Style Transfer}
Style transfer techniques \cite{8732370,cai2023image,li2017demystifying,gatys2016image,zhang2013style,liao2017visual,gu2018arbitrary,zhang2022generalized,kolkin2019style,deng2020arbitrary,yao2019attention,liu2021adaattn,an2021artflow,chen2021artistic,zhang2022domain,wu2021styleformer,deng2022stytr2,cheng2023general,zhang2023inversion,huo2022cast} have made significant advancements in enabling the transfer of artistic styles onto digital images without necessitating a face in the source images. This technology effectively captures and applies the aesthetic elements of a style image across the entirety of a content image. However, for applications like character makeup in visual storytelling, where specific alterations are required without altering the whole facial identity, traditional style transfer proves unsuitable. It applies the style uniformly across the entire image, including facial features, which may result in an overall transformation that diverges from the desired precise makeup application needed for character enhancement.

\begin{figure*}[ht]
	\centering 
	\includegraphics[width=1\textwidth]{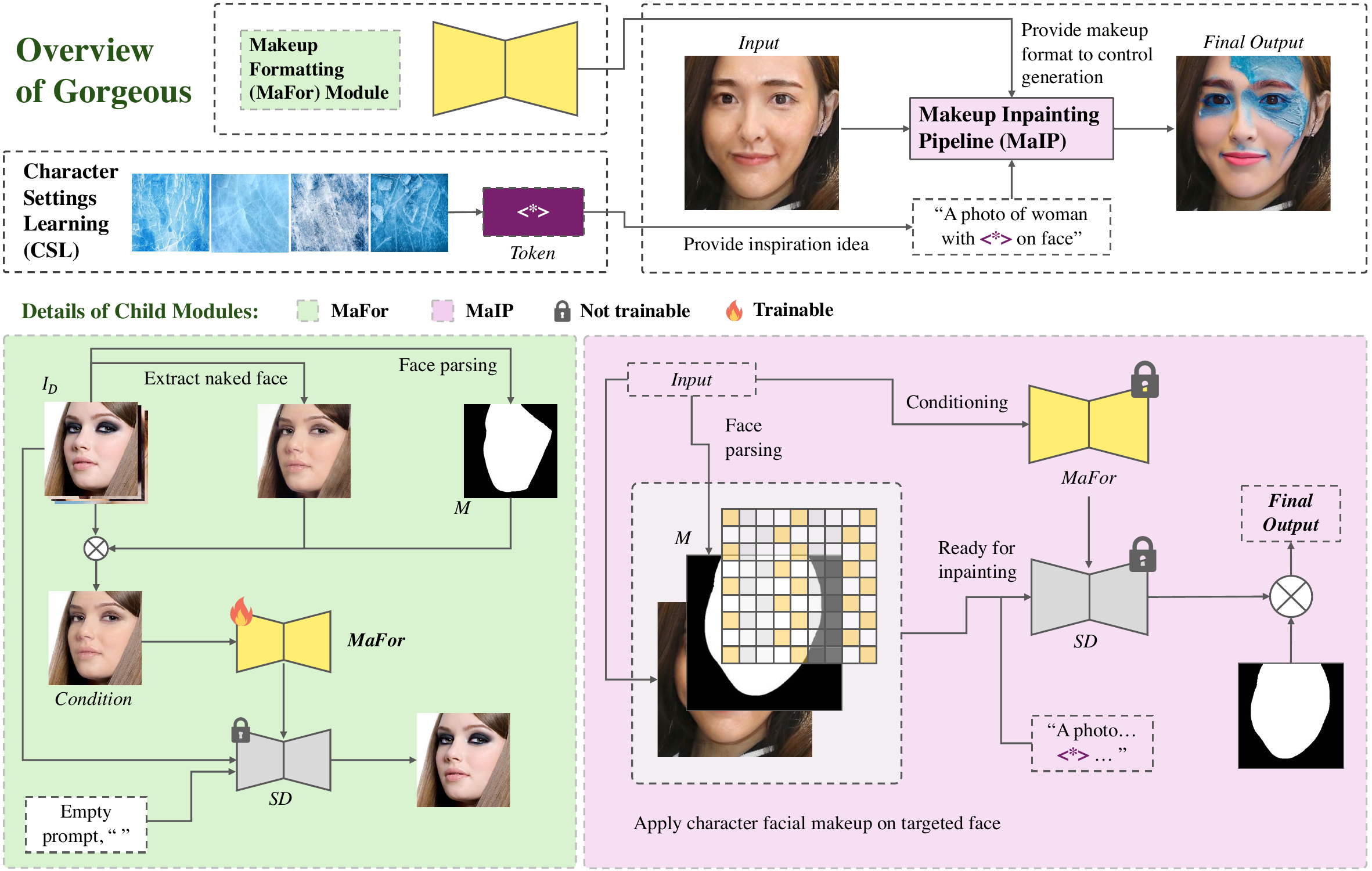}
	\caption{Overall $Gorgeous$ architecture. Given a set of inspirational reference images, these images are processed to extract and embed inspirational elements into a placeholder token. This token is then utilized within the MaIP as a simple textual guide to generate character-specific makeup designs. Throughout this process, the generation is consistently overseen by our pretrained MaFor module, ensuring that the outputs strictly adhere to makeup designs without deviating into unrelated elements.}
	\label{fig:overall_architecture}
\end{figure*}

\subsection{Other Diffusion Models}
Recent diffusion models \cite{ramesh2022hierarchical_dalle,balaji2022ediffi,ding2022cogview2,nichol2021glide,saharia2022photorealistic_imagen,ding2021cogview,Zhang2023AddingCC,Dhariwal2021DiffusionMB,Brooks2022InstructPix2PixLT,Gal2022AnII,Ruiz2022DreamBoothFT,Tumanyan2022PlugandPlayDF,Kawar2022ImagicTR,Kim2021DiffusionCLIPTD,Podell2023SDXLIL,Parmar2023ZeroshotIT} have demonstrated exceptional ability in generating visual content. These include capabilities in text-to-image generation \cite{ramesh2022hierarchical_dalle,balaji2022ediffi,ding2022cogview2,nichol2021glide,saharia2022photorealistic_imagen,ding2021cogview,Zhang2023AddingCC} and text-guided image-to-image translation or editing \cite{Tumanyan2022PlugandPlayDF,Kawar2022ImagicTR,Avrahami2021BlendedDF,nichol2021glide,Brooks2022InstructPix2PixLT,Kawar2022ImagicTR,Kim2021DiffusionCLIPTD,Parmar2023ZeroshotIT,Meng2021SDEditGI}. However, when it comes to generating precise character makeup, the reliance on textual descriptions presents challenges. These descriptions are often too general or lack the necessary specificity, leading to inaccuracies in the desired makeup outcomes. Techniques such as Textual Inversion \cite{Gal2022AnII} and DreamBooth \cite{Ruiz2022DreamBoothFT} attempt to enhance guidance by learning specialized tokens from reference images, yet they struggle to accurately capture and reproduce the intricacies of specific makeup styles, resulting in often undesired results in character makeup generation. 

\subsection{Image Inpainting}
Image inpainting techniques \cite{Rombach_2022_CVPR,Xie2022SmartBrushTA,Yang2023UnipaintAU,Xia2023DiffIRED,fcf,lahiri2020prior,yi2020contextual,zhao2020uctgan,zhou2020learning,liao2020guidance,guo2021image,zeng2022aggregated,zhou2021transfill,zhao2022geofill,bertalmio2000imagediffusion,lugmayr2022repaint,suvorov2022resolution,zhao2021comodgan,nazeri2019edgeconnect,zeng2020profill,li2022mat,li2022misf,liu2022reduceput,zheng2022bridgingtfill,luo2023reference,motamed2023patmat,Ramesh2022HierarchicalTI,Avrahami2021BlendedDF} traditionally fill missing or damaged areas in images using surrounding pixel information, guided by a mask. Adapting this method for digital makeup application, we use the idea of inpainting to seamlessly apply makeup on unadorned faces, leveraging the existing context to ensure natural integration of the makeup with the facial features. This innovative approach not only preserves the integrity of non-facial areas but also enhances the natural appearance of the skin, pushing the boundaries of traditional inpainting into creative cosmetic enhancements.

\section{Methodology}

Our goal is to create a \textit{character facial makeup} for a visual story, where:
\begin{enumerate}[label=(\roman*)]
    \item complex textual descriptions are unnecessary;
    \item the inspiration reference images can be any kind of images, not limited to makeup images; 
    \item the output generated is in the form of makeup while remaining relevant to the reference images; 
\end{enumerate}

As illustrated in Fig. \ref{fig:overall_architecture}, our proposed $Gorgeous$ is a diffusion-based model, which utilizes a pre-trained Stable Diffusion (SD) \cite{Rombach_2022_CVPR} to design its three essential modules: \textbf{(i)} Makeup Formatting (MaFor), a ControlNet-based module that learns to convert any inspirational ideas in the form of text tokens \cite{Gal2022AnII} into a makeup. \textbf{(ii)} Character Settings Learning (CSL) is a module that learns from inspiration sources and encodes the knowledge as a form of text token leveraging textual inversion \cite{Gal2022AnII}. \textbf{(iii)} Makeup Inpainting Pipeline (MaIP) incorporates both MaFor and CSL to generate a seamless character makeup on the face.

To generate a makeup image, we initiate the process with a simple text prompt such as ``\textit{a photo of a woman with <$\ast$> on face}'', where <$\ast$> signifies any text token learned through CSL. Then, the MaFor module uses this prompt, alongside an image of a bare face, to compute the features that guide the UNet of SD during the image generation process. Following this, the MaIP ensures that the desired makeup is confined exclusively to the facial area during the inference stage. Next, we will detail each of the module.

\subsection{Makeup Formatting (MaFor) Module}

The Makeup Formatting (MaFor) module is based on ControlNet \cite{Zhang2023AddingCC} and designed to understand and apply makeup. The term "makeup formatting" describes the module's capability to recognize and execute basic makeup task effectively. This includes tasks such as applying foundation uniformly across the skin, adding blush to the cheeks, and applying eyeliner, eyeshadow, and lipstick to the appropriate facial areas. Technically, MaFor undergoes training on a comprehensive makeup dataset to grasp the essential aspects and techniques of makeup application, thereby enabling it to execute these tasks with both precision and realism which is detailed next.

\paragraph{Paired Data Preparation.} Due to the high costs and scarcity of paired makeup data\footnote{It is challenging to obtain a pair of before and after makeup with aligned face.}, we leverage an unsupervised domain transfer method, LADN \cite{gu2019ladn}, to generate pseudo paired makeup data from unpaired datasets $D$. In short, LADN is employed to effectively transfer a non-makeup facial style from one to a makeup face, i.e., performing a ``demakeup'' process. Let the makeup image as $I_{D} \in \mathbb{R}^{3 \times H \times W}$, we obtain the naked face through LADN as $I_{\text{naked}}^* = \mathrm{LADN}(I)$. This process allows us to simulate a paired dataset where each original makeup face is matched with a corresponding demakeup/naked face. 

Since the domain transfer may unintentionally alter the non-facial regions, a simple trick is to isolate the face region to ensure the accuracy of the makeup application. We utilize an off-the-shelf face parsing module \cite{yu2018bisenet} to accurately segment the face in an image. We then blend $I$ and $I_{\text{naked}}$ to obtain the final naked face as:
\begin{align} \label{eq:1}
    I_{\text{naked}} = I_{\text{naked}}^* \cdot M + I_{D} \cdot (1 - M),
\end{align}
where $M$ is the indicator of facial area in which $M_{ij}=1$ if the $ij$-th patch is the facial region and $M_{ij}=0$ otherwise. In practice, we apply Gaussian blur on $M$ to avoid the sharp edges when blending them.

\paragraph{ControlNet Training.} Employing the pseudo-paired makeup dataset $D_{\text{paired}}$ obtained from the previous step, we train our ControlNet-based MaFor. Specifically, this module is structured to take a naked face $I_{\text{naked}}$ as input to compute the features that guide the final image generation process -- the makeup face:
\begin{align}
    \epsilon_\theta(z_t, p, t, c) = SD(z_t, p, t) + MaFor(z_t, p, t, c),
\end{align}
where $\epsilon_\theta$ is the computed noise, $z_t$ is the noisy latent at timestep $t$, $c$ is the \textit{latent} of naked face which acts as the condition of the controlled generation. Note that we use an empty string (``'') as the text prompt during training to avoid introducing extra information and allow MaFor to directly recognize the semantic relationship between the naked face and the output face (i.e., what is makeup). MaFor is optimized using a diffusion loss function:
\begin{align}
    \mathcal{L}_{\text{control}} = \mathbb{E}_{z,t,\epsilon,c}\left[ \|\epsilon - \epsilon_\theta(z_t,``",t,c)\|^2 \right].
\end{align}

The primary advantage of our ControlNet-based makeup module is its flexibility in getting any desired makeup style through simple text prompts with the correct makeup format and without altering the underlying facial identity. This capability also marks a significant improvement over existing makeup transfer methods \cite{sun2022ssat,yang2022elegant,yan2023beautyrec}, which often rely on more sophisticated pipelines to preserve facial identity during makeup application. 

\subsection{Character Settings Learning (CSL) Module}

While MaFor can apply makeup simply through text prompts (see examples in supplementary material), character makeup should not be constrained by only text prompts, it shall be inspired by arbitrary references, regardless of the presence of a face. To learn makeup from arbitrary styles, we propose a Character Settings Learning (CSL) module.

Inspired by textual inversion \cite{Gal2022AnII}, given a set of 3 to 5 reference images, CSL is designed to learn a specific concept that co-appears in the reference images. Technically, the concept is encoded into a text embedding $v$ that represents the learned makeup style as follows:
\begin{align}
    \mathcal{L}_{ldm} &= \mathbb{E}_{z,t,v,\epsilon}\big[ || \epsilon - SD(z_t,v,t)||^2 \big], \label{eq:diffusionloss} \\
    v^* &= \argmin_v \, \mathcal{L}_{ldm} \label{eq:textinv_obtain},
\end{align}
where $v^*$ is the optimal text embedding. Typical textual inversions use prompts such as ``\textit{a photo of a <$\ast$>}''. Instead, we employ more directed prompts such as ``\textit{a photo of a woman with <$\ast$> on face}''. This guides the learning process to focus on interpreting ``<$\ast$>'' as the makeup style.

Note that CSL operates independently from MaFor. This separation is crucial as CSL is specifically designed to capture inspirational ideas for makeups from arbitrary references without the need for paired data. This flexibility allows CSL to adapt to a wide range of references, making it a potent tool for creative makeup applications that go beyond text-driven methods.

\subsection{Makeup Inpainting Pipeline (MaIP)}

To generate a desired makeup image, $I_{\text{final}}$, we initiate a text prompt $p=$``\textit{a photo of a woman with <$\ast$> on face}'' with <$\ast$> as our desired style which is obtained through CSL. Then, the denoising process is performed as:
\begin{align}
    \epsilon_{\text{uncond}} &=  \epsilon_\theta(z_t,``", t, c), \nonumber \\
    \epsilon_{\text{pred}} &=  \epsilon_\theta(z_t,p, t, c), \nonumber \\
    z_{t-1}^* &= \epsilon_{\text{uncond}} + g \cdot (\epsilon_{\text{pred}} - \epsilon_{\text{uncond}} ), \nonumber \\
    z_{t-1} &= z_{t-1}^* \cdot M^* + c_{t-1} \cdot (1 - M^*) \label{eq:denoise}
\end{align}
where $g$ is guidance scale in classifier-free guidance (CFG) \cite{ho2022cfg}, $p$ is the text prompt, $c$ is the latent of a naked face, $z_{t-1}$ is the noisy latent at timestep $t-1$, $c_{t-1}$ is the noisy latent of the naked face at timestep $t-1$ and $M^*$ is the downsampled version of $M$. The goal of Eq.~\eqref{eq:denoise} is to focus the denoising process exclusively on the facial area.

At timestep $t=0$, the generated latent $z_0$ is obtained and subsequently decoded into the image space, resulting in $I_{\text{gen}}$. Due to quantization errors that occur when decoding from latent to image space, a blending operation is performed on the decoded image to mitigate these artifacts as:
\begin{align}\label{eq:7}
    I_{\text{final}} = I_{\text{gen}} \cdot M + I_\text{naked} \cdot (1 - M),
\end{align}
to ensure the details in the non-facial area are remained.

\begin{figure*}[ht]
	\centering 
	\includegraphics[width=0.8\textwidth]{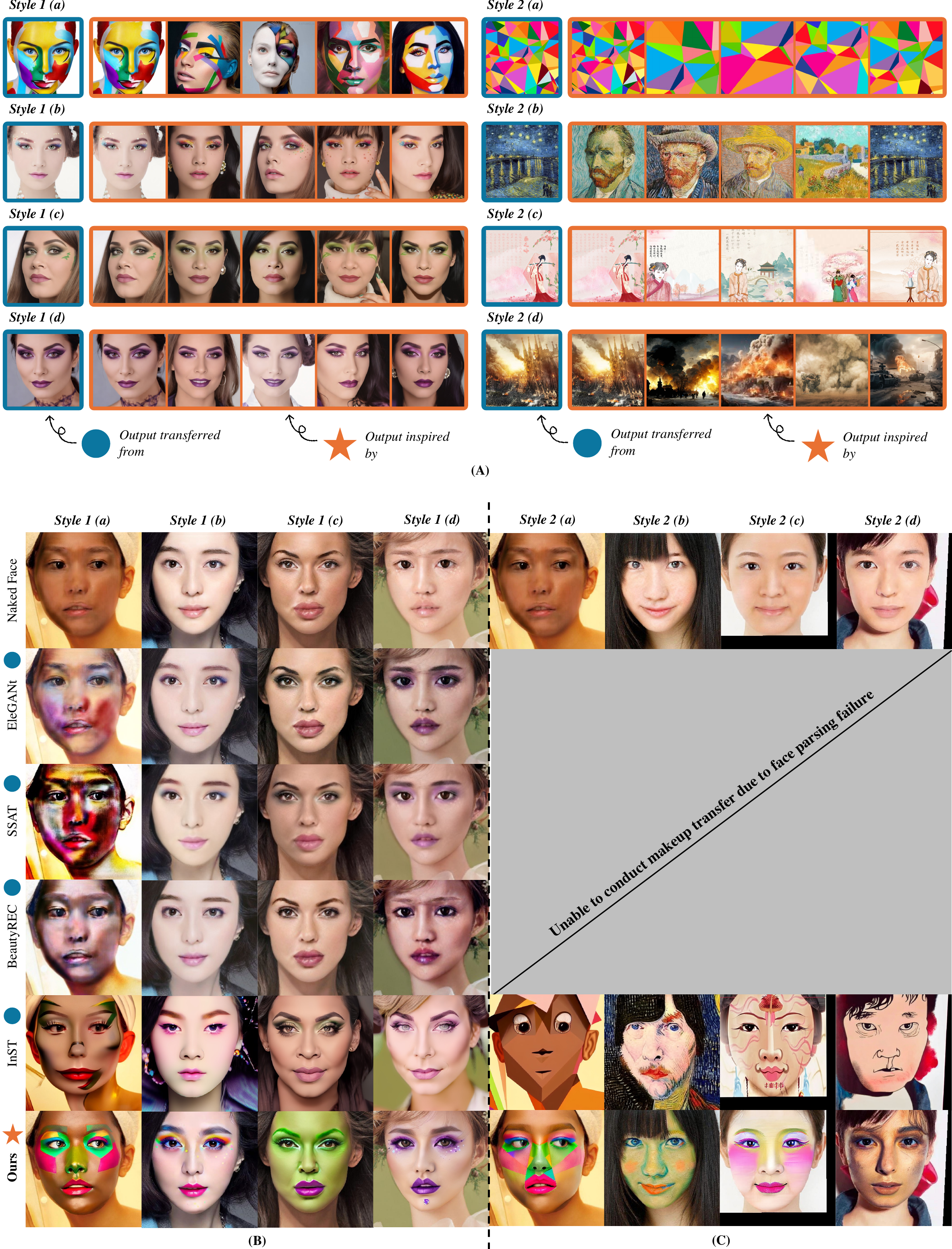}
	\caption{(A) showcases a collection of inspiration reference images for eight distinct character settings. These are divided into two categories: Style 1 (a) to (d), which include reference images featuring human faces, while Style 2 (a) to (d), which comprise images without human faces. Panels (B) and (C) display our qualitative results for Style 1 and Style 2, respectively, in comparison with state-of-the-art makeup transfer methods (i.e., EleGANt, SSAT, BeautyREC) and a style transfer method (i.e., InST). Outputs that directly replicate the style from the reference images are marked with a \textit{blue circle}, while those that are creatively inspired by the styles are indicated with an \textit{orange star}.}
	\label{fig:qual_big}
\end{figure*}

\paragraph{Note.} The design of our generation process is influenced by image inpainting techniques, which aim to "fill in" gaps within a masked image using the surrounding contextual information. To ensure the integrity of the non-masked areas, the resultant image undergoes a blending process. Similarly, we incorporate these principles into our generation process.

Leveraging the self-attention mechanism in the SD model, our approach utilizes all available information, such as facial features, hair, accessories, and background elements as cues to guide the generation of the makeup image. Consequently, we have named our denoising process the Makeup Inpainting Pipeline (MaIP).

\begin{figure*}[th]
	\centering 
	\includegraphics[width=0.65\textwidth]{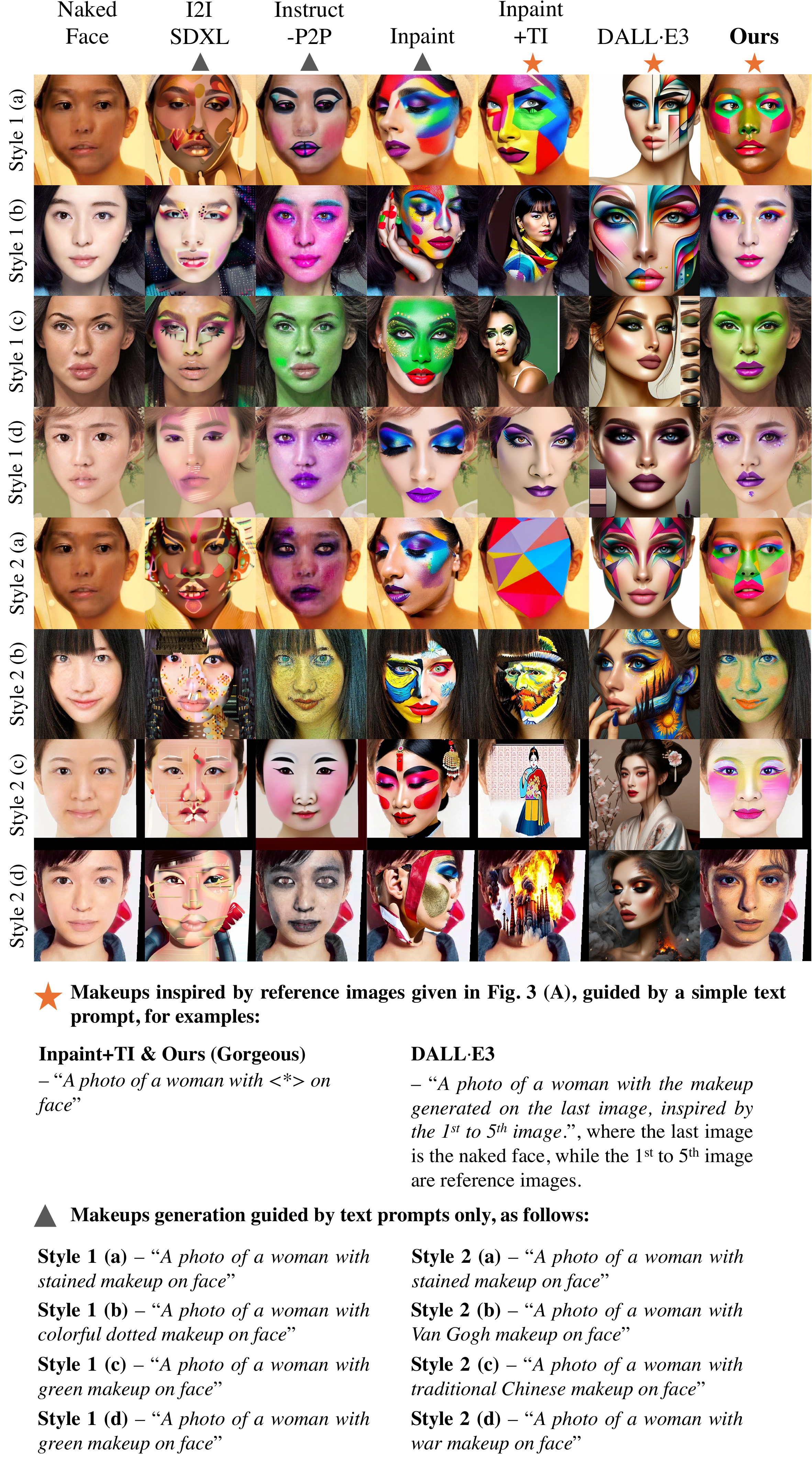}
	\caption{Other qualitative results compared with existing state-of-the-art text-guided image-to-image generation/editing methods to evaluate their capabilities in generating your desired character facial makeups. While the image input is the \textit{Naked Face}, the text prompts are listed in the figure to guide the generation.}
	\label{fig:comparewithothers}
\end{figure*}

\section{Experiments}
\subsection{Datasets and Evaluation Metrics}

\paragraph{Datasets.}
Gorgeous's MaFor module is trained on the BeautyFace \cite{yan2023beautyrec} dataset, which contains 2,447 makeup images. 
To demonstrate the character facial makeup task, we utilize two categories of reference sets: (Style 1) face images with relevant makeup which include images generated from SDXL\footnote{\url{https://huggingface.co/spaces/tonyassi/image-to-image-SDXL}} and images chosen from the website Pinterest\footnote{\url{https://www.pinterest.com/}}; (Style 2) non-facial images with arbitrary styles, for convenience, we called it non-facial style. More image examples are available for Style 1 and 2 in the supplementary material.

\paragraph{Evaluation Metrics.}
To assess the performance of our generated makeups, we employ three key metrics: (i) CSD Similarity \cite{Somepalli2024MeasuringSS} measures how well the generated images emulate the given styles; (ii) DreamSim assesses perceptual similarity, reflecting the human visual perspective to determine how closely the generated images match human expectations; Both CSD and DreamSim are measured with cosine similarity. The reason we chose CSD and DreamSim but not CLIP/DINO score as used in \cite{Ruiz2022DreamBoothFT} is due to CLIP/DINO were trained to understand images semantically and may ignore low-level features such as colors, unlike CSD and DreamSim which were trained to understand style and perceptual similarity. (iii) Fréchet Inception Distance (FID) \cite{heusel2017gans} quantifies the alignment between the generated makeup images and the BeautyFace dataset's makeup format in terms of distribution (i.e., to measure whether the generated images are actually similar to makeup).

\subsection{Baselines} We benchmark Gorgeous against established methods across three domains: \textbf{(1)} Makeup transfer (i.e., EleGANt \cite{yang2022elegant}, SSAT \cite{sun2022ssat}, BeautyREC \cite{yan2023beautyrec}); \textbf{(2)} Style transfer (i.e., InST \cite{zhang2023inversion}); \textbf{(3)} Image-to-image translation/generation (i.e., I2I SDXL \cite{Podell2023SDXLIL}, InstructPix2Pix \cite{Brooks2022InstructPix2PixLT}, Stable Diffusion Inpainting \cite{Rombach_2022_CVPR}, Inpainting with Textual Inversion \cite{Gal2022AnII}\footnote{This means using the token learned through CSL.} and DALL·E3 \cite{dalle3} (OpenAI's ChatGPT4)).

\subsection{Implementation Details}
We utilize a pre-trained SDv2.1 \cite{Rombach_2022_CVPR} as our diffusion model, also for both MaFor (i.e., the ControlNet \cite{Zhang2023AddingCC} is initialized using SDv2.1) and CSL (textual inversion is performed with SDv2.1). All images were resized to $512\times512$. The training for MaFor was conducted with gradient accumulation steps set to 4, a learning rate of 1e-4, and a total of 15,000 training steps with a batch size of 1.
For CSL, the token for each style is trained with a total of 5,000 training steps, with a learning rate of 1e-5 and a batch size of 1.
During inference, the guidance scale $g$ varied from 3 to 20 with the number of inference steps from 30 to 100, depending on the desired makeup intensity. Further details and inference variations are available in the supplementary material.

\begin{table*}[ht]
\centering
\caption{Quantitative evaluation of our method with relevant competitive methods. **The absence of makeup transfer scores for Style 2(a-d) is due to the failure of existing makeup transfer methods to transfer the makeups, given a reference image without a human face. Both CSD and DreamSIM are measured in cosine similarity. Bold value indicates the best score in the column.}
\begin{tabular}{lcccccc}
\toprule
\textbf{Methods} & \multicolumn{3}{c}{\textbf{Average of Style 1(a-d)}} & \multicolumn{3}{c}{\textbf{Average of Style 2(a-d)}} \\
& CSD $\uparrow$ & DreamSIM $\uparrow$ & FID $\downarrow$ & CSD $\uparrow$ & DreamSIM $\uparrow$ & FID $\downarrow$  \\
\midrule
\multicolumn{7}{l}{\textbf{(Makeup transfer)}}\\
EleGANt \cite{yang2022elegant}  & 0.48 & 0.54 & 45.13 & - & - & - \\
SSAT \cite{sun2022ssat}  & 0.45 & 0.56 & 58.92 & - & - & -\\
BeautyREC \cite{yan2023beautyrec} & 0.39 & 0.49 & {\textbf{37.82}} & - & - & -\\
\midrule
\multicolumn{7}{l}{\textbf{(Style transfer)}}\\
InST \cite{zhang2023inversion} & 0.60 & {\textbf{0.67}} & 119.35 &0.29 & 0.29 & 202.00 \\
\midrule
\multicolumn{7}{l}{\textbf{(Image-to-image translation/generation)}}\\
I2I SDXL \cite{Podell2023SDXLIL} & 0.48 & 0.55 & 142.27 &0.15 & 0.22 & 141.75 \\
InstructPix2Pix \cite{Brooks2022InstructPix2PixLT} & 0.51 &0.54 & 94.09 & 0.17 & 0.16 & 101.13\\
Inpainting \cite{Rombach_2022_CVPR} & 0.59 & 0.57 & 146.65 & 0.16 & 0.23 & 169.72\\
Inpainting+TI \cite{Rombach_2022_CVPR,Gal2022AnII}  & 0.58 & {0.63} & 128.37 & \textbf{0.34} & \textbf{0.46} & 250.81\\
\midrule
Ours (Gorgeous) & \textbf{0.61} & 0.61 & {53.29} & 0.21 & 0.28 & \textbf{89.84} \\
\bottomrule
\end{tabular}
\label{tab:quan_table}
\end{table*}

\section{Evaluations}
\subsection{Qualitative Evaluation}

In Fig. \ref{fig:qual_big}(A), Style 1(a) to 1(d) indicate makeup images to be learned or transferred while Style 2(a) to 2(d) indicate non-facial images of different ideas. 

In Fig. ~\ref{fig:qual_big}(B), we visualize the results of makeup transfer/generation. The results show that \textbf{(i)} our \textbf{Gorgeous} generates unique character facial makeups, distinguishing itself from the makeup transfer methods, i.e., EleGANt \cite{yang2022elegant}, SSAT \cite{sun2022ssat}, and BeautyREC \cite{yan2023beautyrec}. \textbf{(ii)}  It is important to note that despite expectations for high fidelity in replicating makeup from reference images to the naked face, traditional makeup transfer methods continue to exhibit significant room for improvement, as they often fail to accurately reproduce the desired styles, especially when the style is exaggerated (e.g., Style 1(a)).

We further visualize our results on transferring makeup styles from non-facial styles in Fig. ~\ref{fig:qual_big}(C). \textbf{(iv)} As existing makeup transfer methods rely heavily on face parsing, they cannot simply transfer makeup style from non-facial styles. \textbf{(v)} Style transfer method (InST) attempts to blend styles globally, they do not specifically adapt the style into makeup formats. \textbf{(vi)} Our Gorgeous successfully transforms stylistic elements from any image into makeup format, demonstrating flexibility and creativity beyond existing makeup transfer and style transfer methods.

As depicted in Fig. \ref{fig:comparewithothers}, further comparisons with other relevant diffusion-based and text-guided image generation models including image-to-image translation with SDXL \cite{Podell2023SDXLIL} (through SDEdit \cite{Meng2021SDEditGI}), InstructPix2Pix \cite{Brooks2022InstructPix2PixLT}, image inpainting with Stable Diffusion \cite{Rombach_2022_CVPR}, the inpainting with Textual Inversion \cite{Rombach_2022_CVPR,Gal2022AnII}\footnote{Note that this means it uses the same token that our Gorgeous learned through CSL.}, and DALL·E3\cite{dalle3}. \textbf{(vii)} Since all these methods are either image generation or image-to-image translation, they struggle to preserve the input face identity. In contrast, Gorgeous excels by accurately producing distinct character looks in makeup form, preserving both the identity and integrity of the original face. This streamlined approach highlights our method’s superior ability to handle diverse image types and complex makeup challenges, establishing a new standard for character facial makeup generation.

\subsection{Quantitative Evaluation}

In Tab.~\ref{tab:quan_table}, we summarize the scores for all methods. \textbf{(i) Gorgeous:} Our method outperforms other image generation models with the lowest FID scores (53.29 for Style 1 and 89.84 for Style 2), indicating that Gorgeous can translate inspiration into makeup formats that align with the BeautyFace dataset better than various baselines. Notably, Gorgeous also performs comparably well as shown by CSD and DreamSim, indicating its effectiveness in matching and perceiving style relevance.
\textbf{(ii) Makeup Transfer Methods:}  In Style 1 (a-d), while BeautyREC \cite{yan2023beautyrec} (37.82) and EleGANt \cite{yang2022elegant} (45.13) show low FID scores, they lack diversity and uniqueness in makeup generation, primarily replicating existing makeups rather than creating new ones. These scores serve as benchmarks but suggest that significant advancements are still needed in traditional makeup transfer methods. While in Style 2 (a-d), which involves reference images without faces, existing makeup transfer methods could not be evaluated due to their high reliance on the face parsing module, highlighting a significant limitation in flexibility.
\textbf{(iii) Style Transfer Performance:}  InST \cite{zhang2023inversion} demonstrates high similarity scores (DreamSIM: 0.67 for Style 1, 0.29 for Style 2; CSD: 0.60 for Style 1, 0.29 for Style 2), yet it underperforms in FID (119.35 for Style 1, 202.00 for Style 2), indicating a divergence from the standard makeup format (i.e., it transfers the whole image, which instead of applying makeup). 
\textbf{(iv) I2I translation/generation:} Inpainting + TI surprisingly scores well in DreamSIM (0.63 for Style 1, 0.46 for Style 2) and CSD (0.34 for Style 2), but it shows poor FID results (128.37 for Style 1, 250.81 for Style 2). In other words, it captures the style from reference images, but it fails to preserve the face identity and does not accurately render the style as wearable makeup.

In summary, Gorgeous performs comparably well across all metrics, particularly in creating character looks that closely match the predefined makeup format of the BeautyFace dataset.

\begin{table}[t]
\centering
\caption{User study: 100 participants are asked to vote for their preferred character makeup designs, which are based on a set of provided reference images.}
\setlength\tabcolsep{4pt}
\begin{tabular}{lcccccccc}
\toprule
\textbf{Methods} & \multicolumn{4}{c}{\textbf{Style 1} (\%) $\uparrow$} & \multicolumn{4}{c}{\textbf{Style 2} (\%) $\uparrow$}\\
 & (a) & (b) & (c) & (d) & (a) & (b) & (c) & (d)\\
\midrule
EleGANt  & 1& 12 & \textbf{38}& 26 &0 &0 &0 &0 \\
SSAT  &1 & 3& 9 & 17 & 0 &0 &0&  0  \\
BeautyREC &1 & 2&1 & 5& 0 &0 &  0 & 0 \\
InST &0 & 4 & 9 & 0 & 2 &1 &6  &0  \\
I2I SDXL& 0 &0 & 1 & 0 & 2& 2 & 2 & 0  \\
InstructPix2Pix & 0 &  0&1 & 5 & 1 & 12 &2 & 18 \\
Inpaint &5 &0 & 0 & 0 & 0 & 4 &4 & 0  \\
Inpaint+TI& 6 & 0& 3 & 0 & 0& 1 & 0& 2   \\
DALL·E3 & 14 & 1 & 8 & 4 & 15 & 9 & 17 &11 \\
\hline
Ours (Gorgeous) & \textbf{72} & \textbf{78} & 30 & \textbf{43} & \textbf{80}  & \textbf{71} &  \textbf{69} &  \textbf{69}  \\
\bottomrule
\end{tabular}
\label{tab:user_study}
\end{table}

\begin{figure*}[ht]
	\centering 
	\includegraphics[width=1\textwidth]{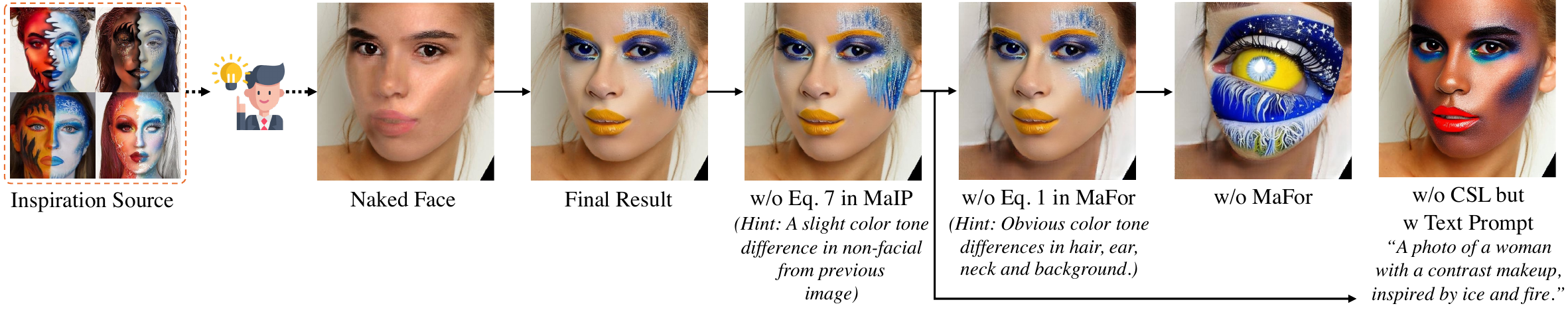}
	\caption{Ablation Study}
	\label{fig:ablation_study}
\end{figure*}

\subsubsection{User Study}

It is important to emphasize that the scores mentioned earlier (i.e., CSD, DreamSIM and FID) serve merely as reference as preferences for character makeup are highly subjective. To address this, we conducted a user study with 100 participants. These individuals reviewed the character makeups displayed in Fig. \ref{fig:qual_big} and Fig.\ref{fig:comparewithothers}. Each participant was shown the bare face along with the inspirational reference images and asked to vote for the makeup they found most appealing and relevant to the reference images.

As presented in Tab. \ref{tab:user_study}, the makeup designs generated by our method, $Gorgeous$, received the highest number of votes in most styles. This was particularly evident in Style 1(a), 1(b), 1(d), and comprehensively in Style 2 (a-d), demonstrating a strong preference (over 70\% of votes) for our method's. In Style 1(c), our makeups design was ranked second, slightly behind EleGANt, which secured 30\% of the votes. These results clearly demonstrate that $Gorgeous$ consistently outperforms competing methods in meeting user preferences, affirming its effectiveness in creating visually appealing and relevant character makeup as evaluated by actual users.

\section{Ablation Study}

In Fig. \ref{fig:ablation_study}, we present a qualitative evaluation of the impact of different modules on the generation of character makeups. The results of removing specific components illustrate their significance in maintaining the integrity and accuracy of the makeup generation process: \textbf{(i) Eq.~\eqref{eq:7} in MaIP:} Omitting Eq.~\eqref{eq:7} results in slight deviations in non-facial areas. While these changes may appear minor, they are critical during inference, as non-target areas should remain unchanged from the input to ensure consistency. \textbf{(ii) Eq.~\eqref{eq:1} in MaFor:} The removal of Eq.~\eqref{eq:1} has a noticeable impact, leading to significant differences in non-facial areas compared to the original naked face. This demonstrates the importance of MaFor in maintaining the overall balance and integrity of the image. \textbf{(iii) MaFor Removal:} Eliminating the MaFor module, which converts the inspirational ideas to makeup format, results in outputs that neither represent makeup nor retain the face identity. This underscores MaFor's crucial role in defining and applying the makeup style correctly. \textbf{(iv) CSL Module:} Without the CSL module to derive ideas from inspirational sources, the model struggles to incorporate thematic concepts such as ice and fire into the makeup. This highlights the module's role in translating complex thematic elements from the inspiration images into the final makeup.

Each component's removal distinctly affects the system’s performance, underscoring their collective importance in achieving high-quality, thematic makeup generation that aligns with user expectations and design intent.

\section{Conclusion}
We introduced \textit{Gorgeous}, an innovative makeup application method specifically designed to generate character facial makeups for visual storytelling. Unlike traditional makeup transfer techniques, which merely replicate makeup from a source face to a target face, \textit{Gorgeous} innovates by enabling the creation of unique and diverse character makeups. Crucially, our method does not rely solely on reference images that contain faces with makeup; instead, it can draw inspiration from a wide range of themed images, even those without visible makeup. This flexibility allows for a broader and more creative approach, making it possible for users to seamlessly translate any thematic inspiration into relevant character makeups.  This is achieved through our innovative creation of multiple components, including the MaFor module for converting the inspiration elements into makeup format, the CSL module for interpreting and incorporating thematic elements from non-facial reference images, and importantly, the MaIP pipeline to apply these thematic makeups precisely onto the face, ensuring that the original identity is preserved while embodying the intended artistic concepts. Extensive experiments confirm $Gorgeous$'s superior performance in both qualitative and quantitative assessments, showcasing its ability to consistently produce high-quality, accurate makeups desgin that enhance visual storytelling. It is our believe that $Gorgeous$ will significantly benefit multimedia industries involved in visual narratives by providing artists and creators with powerful tools to craft unique, engaging, and immersive character makeups.

\bibliographystyle{ieeenat_fullname}
\bibliography{arxiv}

\appendix
\clearpage

\section*{Supplementary Material}

\section{Introduction}
In this supplementary material, we provide more detailed on implementation insights and comprehensive experimental results to underscore the potential and efficacy of our new makeup application method, $Gorgeous$. This novel approach surpasses traditional makeup application techniques by enabling the creation of unique and creatively inspired facial makeups from \textit{any thematic idea, marking a first in the field}. Here, we elaborate on:
\begin{enumerate}[label=(\roman*)]
    \item The overarching potential of $Gorgeous$, highlighting its advancements over existing methods.
    \item A thorough comparison of $Gorgeous$ with other makeup, style transfer and relevant generation methodologies.
    \item Technical details for MaFor.
    \item The limitation, and future direction of our research.
    \item Extensive datasets for Style 1 and Style 2.
    \item Additional qualitative and quantitative results that showcase the capabilities of $Gorgeous$ across varied scenarios. 
    \item Further analysis on varied implementation details.
    \item Additional ablation study.
\end{enumerate}

\section{Potential of $Gorgeous$, a brand new makeup application method}

Current research on makeup application primarily focuses on makeup transfer techniques, which aim to ``replicate'' makeup styles from one face to another. However, this approach falls short in fostering creativity and diversity in makeup designs. To address this significant gap, we introduce $Gorgeous$—the \textbf{first work} in makeup realm designed to generate unique and diverse makeup styles inspired by \textit{any conceptual} idea. 

\subsection{Advantages of $Gorgeous$ over traditional makeup application methods:}
\begin{enumerate}[label=(\roman*)]
   \item {\textbf{Creativity and Uniqueness:}} $Gorgeous$ enables the creation of unique and distinct makeup styles that break free from the limitations of traditional transfer methods.
    \item {\textbf{Flexibility:}} It supports flexible makeup generation from a wide array of inspirations, not confined to existing facial makeup designs.
    \item {\textbf{Reference-Independence:}} Unlike conventional methods, $Gorgeous$ does not require a face in the reference images, thus sidestepping common issues associated with makeup transfer.
    \item {\textbf{Seamless Makeup Application While Maintaining Target Image's Integrity:}} It ensures seamless makeup application while maintaining the subject's facial identity and the integrity of non-facial regions.
\end{enumerate}

\textbf{Broad Applications of $Gorgeous$:}
\begin{enumerate}[label=(\roman*)]
    \item \textbf{Entertainment and Media:} Ideal for creating character-specific facial makeups for use in films, television, theater, cosplay events, and fashion shows, as detailed in the main document.
    \item \textbf{Everyday Use:} Facilitates the digital application of everyday makeup on photographs, enhancing personal aesthetics with ease.
    \item \textbf{Virtual Try-Ons:} Traditional makeup transfer methods, initially developed for virtual try-ons, typically necessitate the preparation of multiple makeup samples to preview different looks on a target face. This approach becomes cumbersome and wasteful, particularly when sample images do not match the user's preferences, requiring additional makeup application and removal to test each new style physically. $Gorgeous$ overcomes these inefficiencies by enabling users to digitally preview a diverse array of makeup styles with ease. This digital method allows for rapid, cost-effective iterations over various styles, ensuring users can find their preferred look without the need for excessive physical trials or the wasteful use of cosmetics.
\end{enumerate}

\section{Thorough comparison of $Gorgeous$ with Other Relevant Methods}
\begin{table*}[ht]
\centering
\caption{Comparative overview of our method $Gorgeous$ with existing possible makeup application methods.}
\
\begin{tabular}{lcccccccc}
\toprule
\textbf{Methods} & Makeup  & Style  & I2I  & InstructP2P & Inp & Inp+TI & DALL·E3 & \textbf{Ours}  \\
 &  Transfer &  Transfer &  SDXL &  &  &  &  &   \\
\midrule
Adds generic contents on face & \cmark & \cmark & \cmark & \cmark & \cmark & \cmark & \cmark & \cmark \\
Applies specific makeup & \cmark & \xmark & \xmark & \xmark & \cmark & \xmark & \cmark & \cmark \\
Creates diverse content  & \xmark & \xmark & \cmark & \cmark & \cmark & \cmark & \cmark & \cmark  \\
References w/o faces & \xmark & \cmark & \cmark & \cmark & \cmark & \cmark & \cmark & \cmark \\
No facial annotations  & \xmark & \cmark & \cmark & \cmark & \cmark & \cmark & \cmark & \cmark \\
Supports textual cues & \xmark & \xmark & \cmark & \cmark & \cmark & \cmark &  \cmark& \cmark\\
Utilizes image references & \cmark & \cmark & \xmark & \xmark & \xmark & \cmark & \cmark & \cmark\\
Preserves face identity & \cmark & \xmark & \xmark & \cmark & \xmark & \xmark & \xmark & \cmark\\
Preserves non-facial area's integrity & \cmark& \xmark & \xmark & \xmark & \cmark & \cmark & \xmark & \cmark\\

\bottomrule
\end{tabular}
\label{tab:comparison_table}
\textit{Note: ``Adds generic contents on face'' refers to any additions applied to the face, while ``Applies specific makeup'' refers to the contents added on face are makeups. ``References w/o faces'' indicates that facial images are not necessary in reference images to inspire the makeup generations.}
\end{table*}

In Tab. \ref{tab:comparison_table}, we present a comprehensive comparison of our method against various state-of-the-art techniques for makeup applications. The methods compared include traditional Makeup Transfer (i.e., EleGANt \cite{yang2022elegant}, SSAT \cite{sun2022ssat}, and BeautyREC \cite{yan2023beautyrec}, Style Transfer \cite{zhang2023inversion}, Image-to-Image Synthesis using Stable Diffusion XL (I2I SDXL) \cite{Podell2023SDXLIL}, Instruct Pixel-to-Pixel (InstructP2P) \cite{Brooks2022InstructPix2PixLT}, and advanced generative models such as DALL·E 3 \cite{dalle3}. We also consider incremental improvements in these areas, such as Inpainting (Inp) \cite{Rombach_2022_CVPR} and Inpainting with Textual Inversion (Inp+TI) \cite{Gal2022AnII}.

Our method excels in several key areas, significantly outperforming other methods in preserving both facial identity and the integrity of non-facial areas. Moreover, it supports both textual and image-based inputs for makeup references, accommodating a wider range of creative possibilities. This flexibility, combined with our method's ability to handle makeup applications without the need for facial features annotations or even the presence of a face in the reference images, sets our approach apart as a highly versatile and effective, new solution for makeup applications.

\section{MaFor's Implementation}

Our MaFor module is structured to take a naked face $I_{\text{naked}}$ as input to compute the features that guide the final image generation process -- the makeup face:
\begin{align}
    \epsilon_\theta(z_t, p, t, c) = SD(z_t, p, t) + MaFor(z_t, p, t, c),
\end{align}
where $\epsilon_\theta$ is the computed noise, $z_t$ is the noisy latent at timestep $t$, $c$ is the \textit{latent} of naked face which acts as the condition of the controlled generation.

$MaFor(\cdot)$ is a trainable variant of $SD(\cdot)$ with its own parameters. The outputs of $MaFor(\cdot)$ are integrated into the middle and upsample blocks of $SD(\cdot)$ respectively. Notably, the latent representation $c$ is dimensioned at $64\times64$, aligning with the size of $z_t$. The original image $I_{\text{naked}}$ undergoes processing through a series of learnable convolutional layers, which downsample it by a factor of $8\times$ to generate the latent $c$. For more details, we refer readers to the original paper \cite{Zhang2023AddingCC}. 
\begin{figure*}[ht]
    \centering 
    \includegraphics[width=0.75\linewidth]{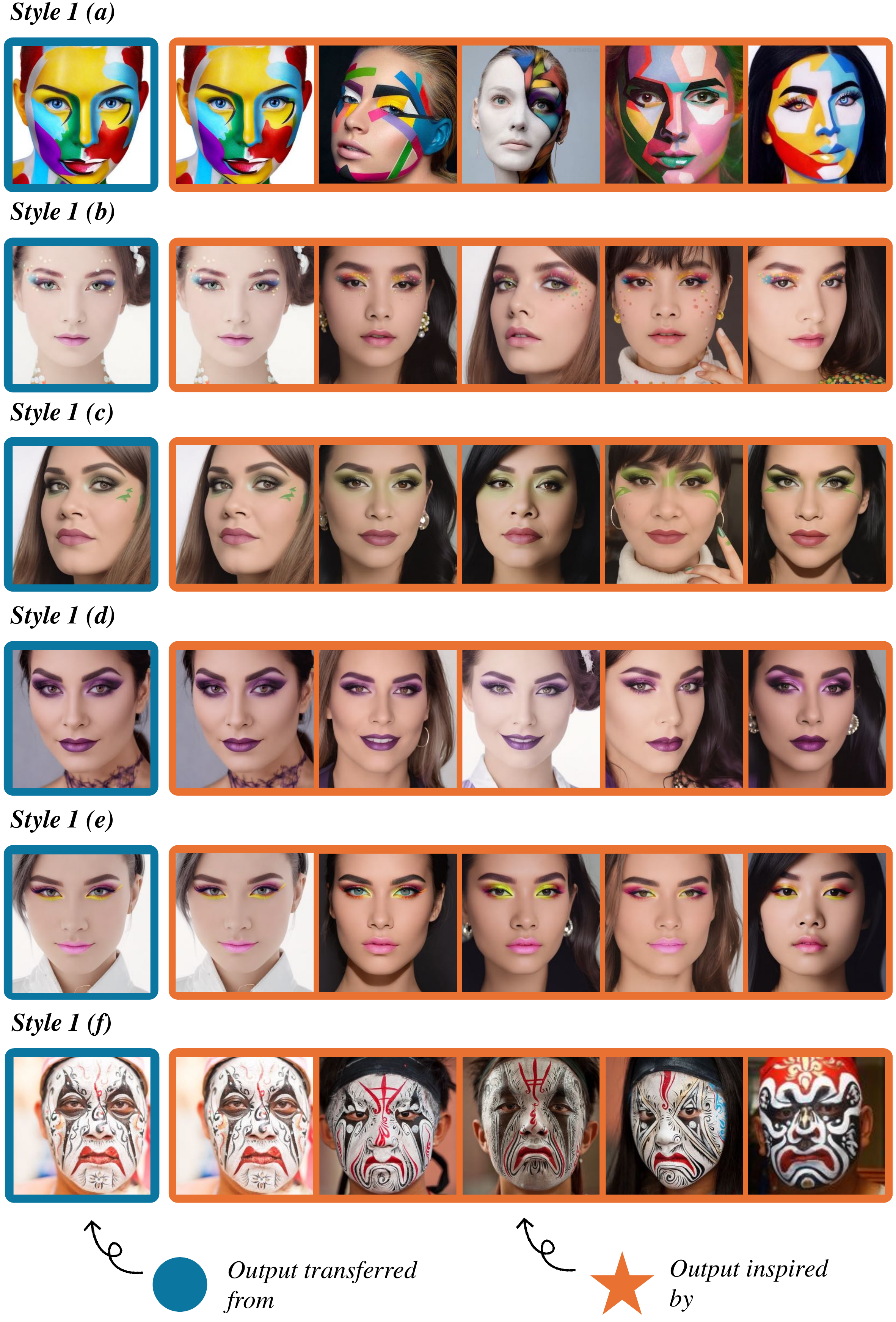}
    \caption{Collection of Facial Makeup Images (Style 1): This figure showcases a variety of makeup styles that have either been generated through SDXL \cite{Podell2023SDXLIL} or curated from Pinterest, illustrating the diversity of facial aesthetics that $Gorgeous$ and other relevant methods can inspire or transfer from. \textit{Note:} The \textit{blue circle} indicates the image selected for makeup or style transfer applications. Images marked with an \textit{orange star} are used as inspiration for the generative processes.
    }
    \label{fig:style1_bank}
\end{figure*}



\section{Limitations and Future Directions}

Current evaluation methodologies, including DreamSIM \cite{Fu2023DreamSimLN}, CSD \cite{Somepalli2024MeasuringSS}, and FID \cite{heusel2017gans}, predominantly assess style, perceptual similarity, and the statistical distances between datasets. However, these metrics are not tailored specifically for makeup assessments—they provide a global style evaluation rather than focusing on the nuances that are crucial for facial makeup analysis. This general approach can overlook the subtle yet critical aspects unique to makeup application, such as color accuracy, textural fidelity, and the integration of makeup with facial features.

\textbf{Future Directions:}
To address this gap, our future work will focus on developing new metrics specifically designed to evaluate makeup style similarity. These metrics will aim to capture the intricacies of makeup application on the face, considering factors like:
\begin{itemize}
    \item \textbf{Color Harmony:} Assessing how well the makeup colors integrate with the natural skin tones and other facial features.
    \item \textbf{Textural Alignment:} Evaluating the realism and appropriateness of makeup textures, ensuring they blend seamlessly without appearing overimposed or unnatural.
    \item \textbf{Contextual Relevance:} Ensuring that the makeup style aligns with the intended theme or inspiration, enhancing the overall look without overpowering the wearer's natural features.
\end{itemize}

These advancements in evaluation metrics will not only enhance the accuracy of makeup assessments but will also contribute to the broader field of aesthetic evaluations in digital and augmented reality applications, paving the way for more personalized and context-aware makeup recommendations and applications.

\section{Datasets: Collections of Inspiration Reference Images}
To effectively demonstrate the capabilities of our facial makeup generation method, we employ two distinct categories of reference sets. These sets are designed to showcase the versatility of $Gorgeous$ in creating character facial makeups under varied thematic influences:
\subsection{Style 1: Facial Images with Makeup}
For character settings closely aligned with traditional makeup styles, we use:
\begin{enumerate}[label=(\roman*)]
    \item \textbf{Images Generated from SDXL (Style 1(b) to 1(e) in Fig. \ref{fig:style1_bank}):} We input specific text prompts describing the desired makeup into the SDXL system to generate images. This method, while innovative, has proven to be somewhat cumbersome and at times unreliable for capturing the exact desired makeup effects.
    \item \textbf{Curated Images from Pinterest (Style 1(a) and 1(f) in Fig. \ref{fig:style1_bank}):} Selected makeup images from Pinterest (\url{https://www.pinterest.com/}) demonstrate the ability of $Gorgeous$ to adapt and recreate styles found in wildly varying online datasets. These images provide a rich source of real-world makeup styles that enhance the diversity and applicability of our approach.
\end{enumerate}

\subsection{Style 2: Non-facial Images with Arbitrary Styles}
\begin{figure*}[t]
    \centering 
    \includegraphics[width=1\textwidth]{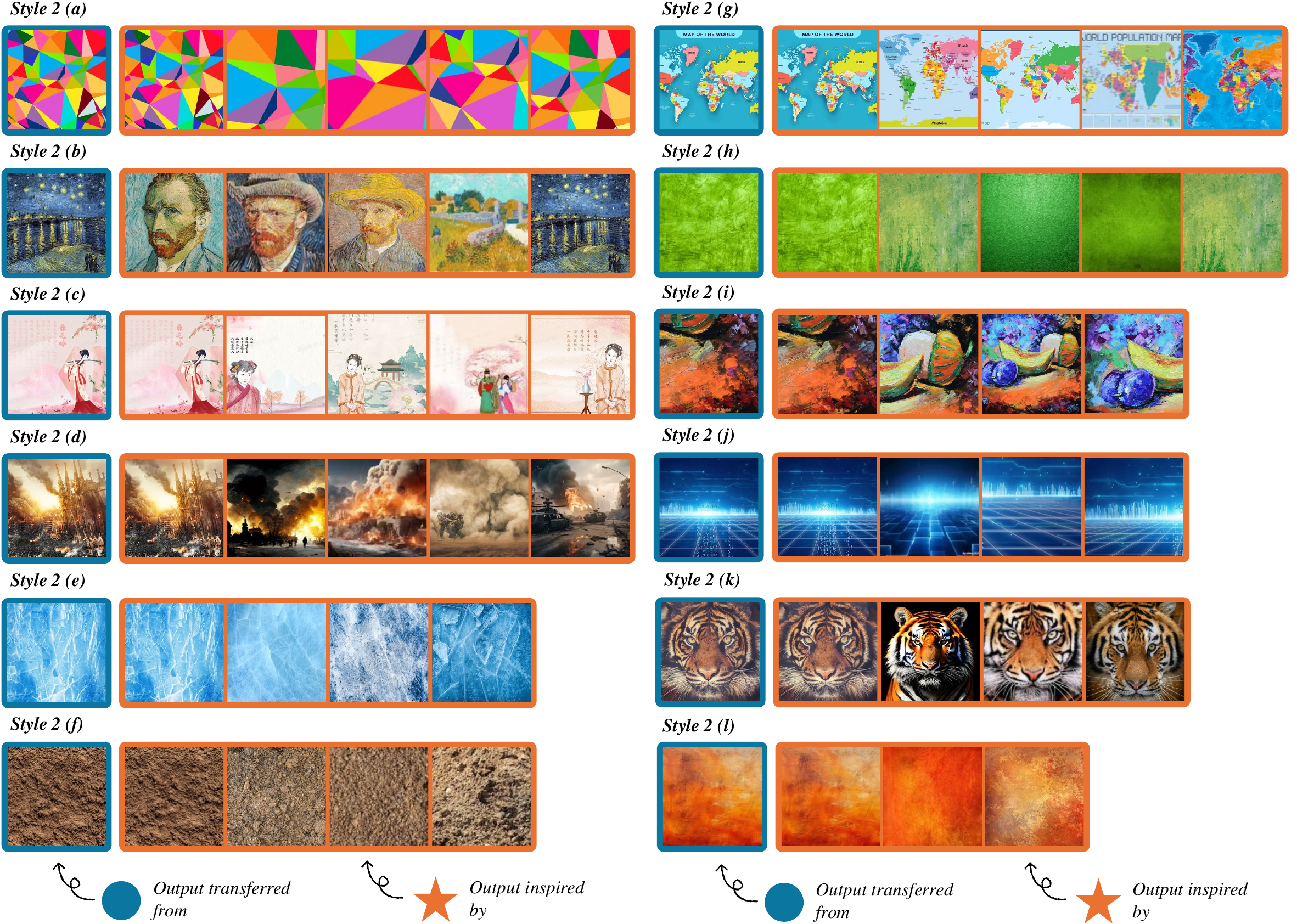}
    \caption{Collection of Non-facial Inspirational Images (Style 2): Displaying a range of abstract and non-traditional inspirations for makeup styles, this figure illustrates how $Gorgeous$ can derive aesthetic cues from non-facial sources, extending the creative possibilities of makeup application. \textit{Note:} The \textit{blue circle} indicates the image selected for makeup or style transfer applications. Images marked with an \textit{orange star} are used as inspiration for the generative processes.
}
    \label{fig:style2_bank}
\end{figure*}

Recognizing that inspiration for makeup can come from non-traditional sources, we also explore:

\begin{enumerate}[label=(\roman*)]
    \item \textbf{Arbitrary Style Inspirations:} This category comprises images without faces, collected randomly from various online platforms:
    \begin{enumerate}
        \item{\url{https://616pic.com/}}
        \item{\url{https://zh.lovepik.com/}}
        \item{\url{https://stock.adobe.com/}}
        \item{\url{https://www.123rf.com/}}
        \item{\url{https://www.vecteezy.com/}}
        \item{\url{https://www.dreamstime.com/}}
        \item{\url{https://www.rawpixel.com/}}
        \item{\url{https://wallpapers.com/}}
        \item{\url{https://depositphotos.com/}}
        \item{\url{https://www.shutterstock.com/}}
        \item{\url{https://www.highmowingseeds.com/}}
        \item{\url{https://www.hgtv.com/}}
        \item{\url{https://www.gardendesign.com/}}
        \item{\url{https://www.ikea.com/}}
        \item{\url{https://laidbackgardener.blog/}}
        \item{\url{https://www.nga.gov/}}
        \item{\url{https://www.metmuseum.org/}}
        \item{\url{https://www.britannica.com/}}
        \item{\url{https://www.vogue.fr/}}
        \item{\url{https://en.wikipedia.org/}}
        \item{\url{https://populationeducation.org/}}
        \item{\url{https://originalmap.co.uk/}}
        \item{\url{https://printmeposter.com/}}
        \item{\url{https://www.tiverton.bham.sch.uk/}}
        \item{\url{https://www.istockphoto.com/}}
    \end{enumerate}
    As shown in Fig. \ref{fig:style2_bank}. These images serve as a basis for demonstrating $Gorgeous$'s ability to conceptualize and implement makeup styles inspired by non-facial elements, showcasing the model's adaptability to abstract inspirations.
\end{enumerate}

\newpage
\section{Additional Qualitative and Quantitative Evaluations}
\subsection{Extended Qualitative Evaluations for Style 1 and Style 2} 
\begin{figure}[h]
	\centering 
	\includegraphics[width=0.5\textwidth]{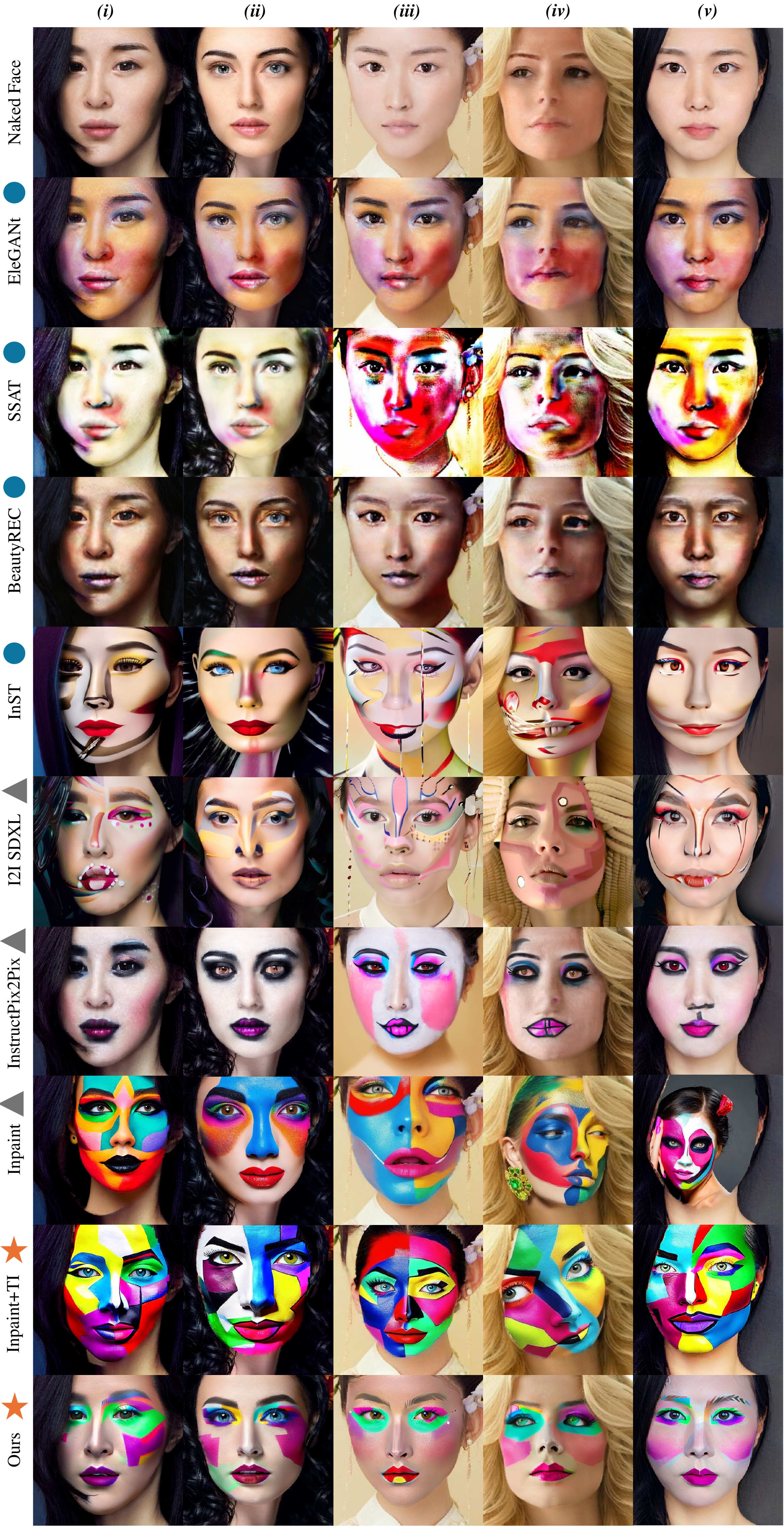}
	\caption{Comparative qualitative results for Style 1 (a), showcasing the performance of our method against other baseline approaches. \textit{Note:} The \textit{blue circle} signifies that the method utilized the specific image referenced by the blue circle in Fig. \ref{fig:style1_bank}. Methods annotated with an \textit{orange star} have drawn inspiration from the images indicated by the orange star in Fig. \ref{fig:style1_bank}. In contrast, a \textit{gray triangle} denotes methods that rely solely on a textual description, specifically the prompt ``A photo of a woman with art makeup on face,'' to guide the makeup generation process.}
	\label{fig:style1a}
\end{figure}

\begin{figure}[ht]
	\centering 
	\includegraphics[width=0.5\textwidth]{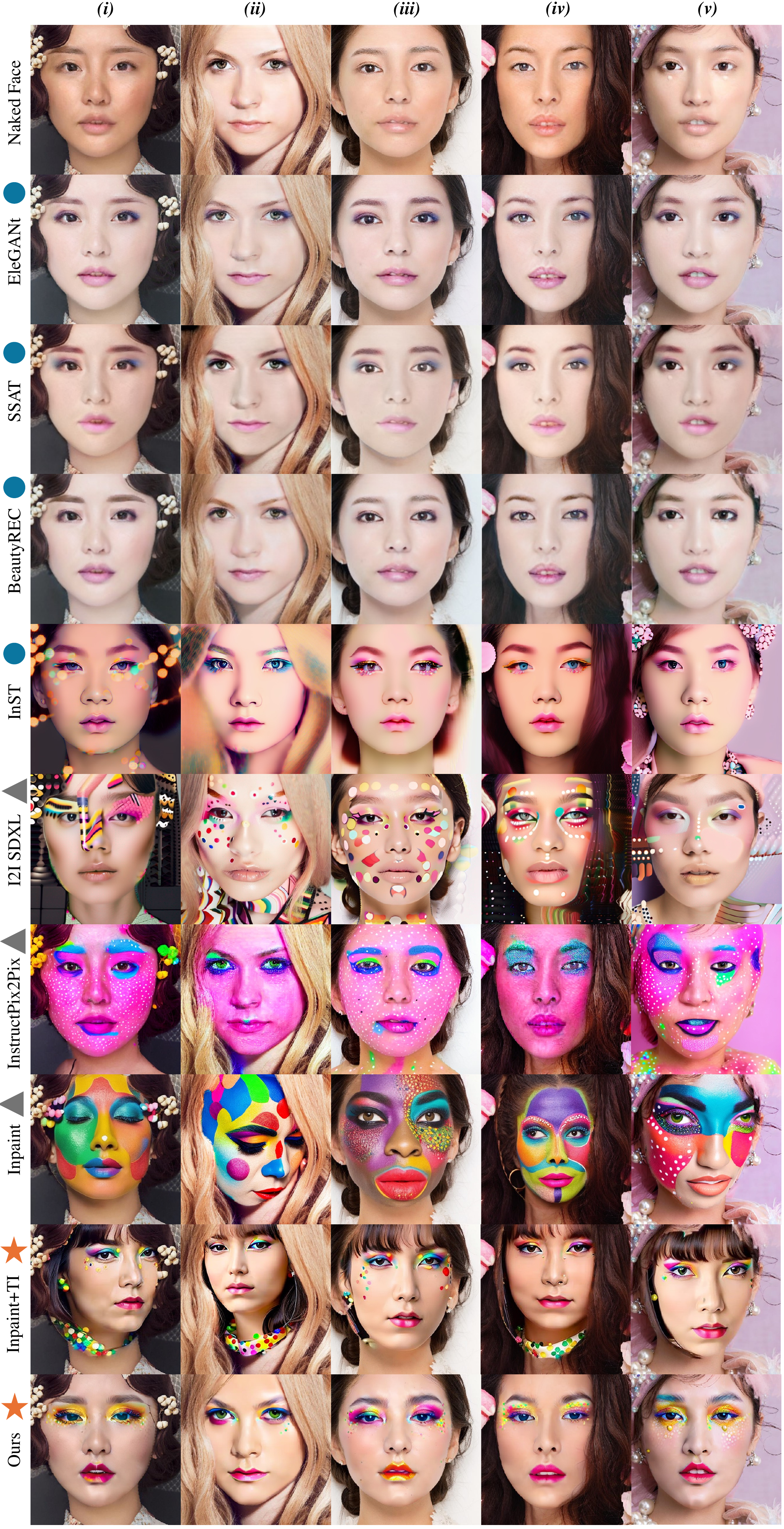}
	\caption{Comparative qualitative results for Style 1 (b), showcasing the performance of our method against other baseline approaches. \textit{Note:} The \textit{blue circle} signifies that the method utilized the specific image referenced by the blue circle in Fig. \ref{fig:style1_bank}. Methods annotated with an \textit{orange star} have drawn inspiration from the images indicated by the orange star in Fig. \ref{fig:style1_bank}. In contrast, a \textit{gray triangle} denotes methods that rely solely on a textual description, specifically the prompt ``A photo of a woman with colorful-dotted makeup on face,'' to guide the makeup generation process.}
	\label{fig:style1b}
\end{figure}

\begin{figure}[ht]
	\centering 
	\includegraphics[width=0.5\textwidth]{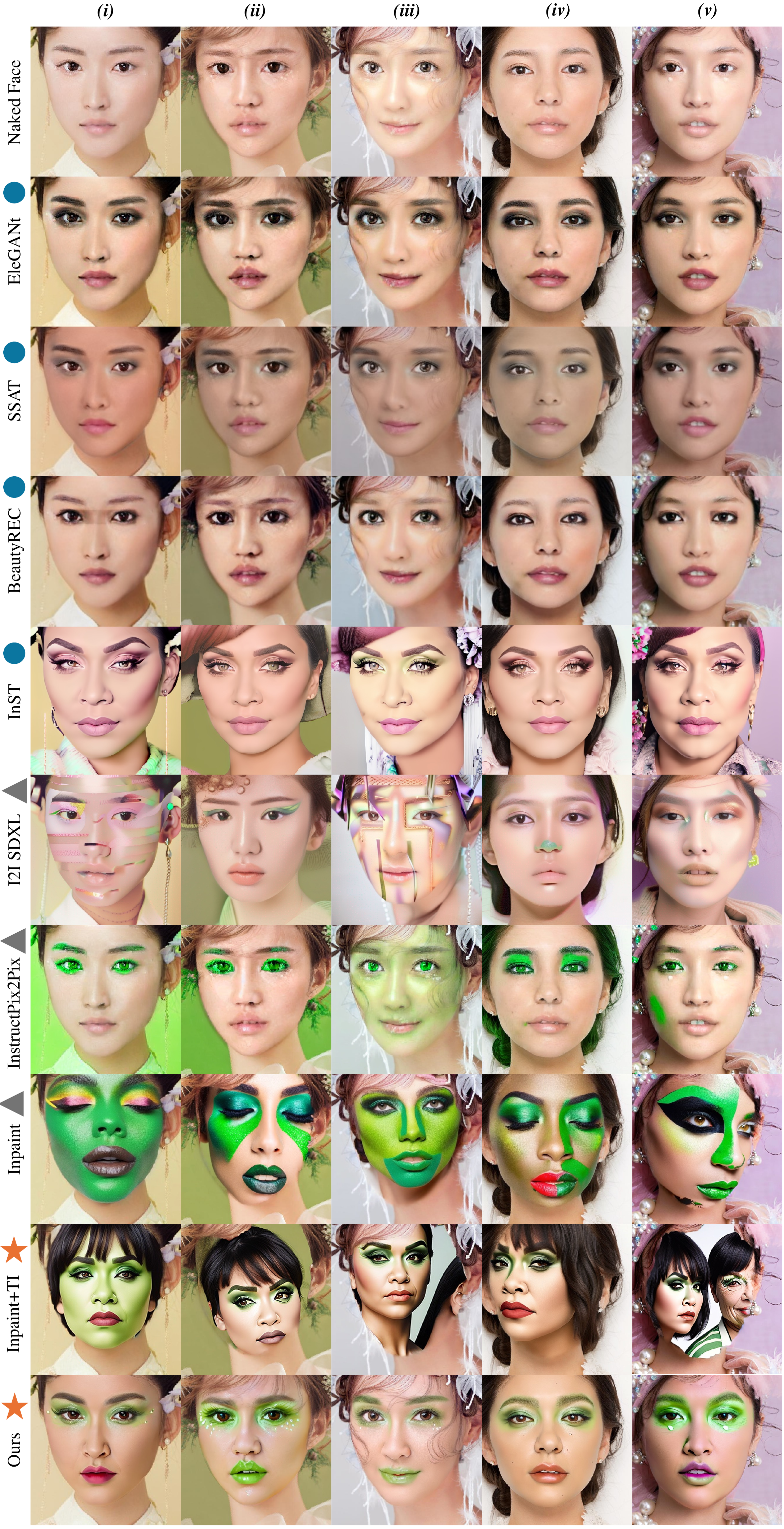}
	\caption{Comparative qualitative results for Style 1 (c), showcasing the performance of our method against other baseline approaches. \textit{Note:} The \textit{blue circle} signifies that the method utilized the specific image referenced by the blue circle in Fig. \ref{fig:style1_bank}. Methods annotated with an \textit{orange star} have drawn inspiration from the images indicated by the orange star in Fig. \ref{fig:style1_bank}. In contrast, a \textit{gray triangle} denotes methods that rely solely on a textual description, specifically the prompt ``A photo of a woman with green makeup on face,'' to guide the makeup generation process.}
	\label{fig:style1c}
\end{figure}

\begin{figure}[ht]
	\centering 
	\includegraphics[width=0.5\textwidth]{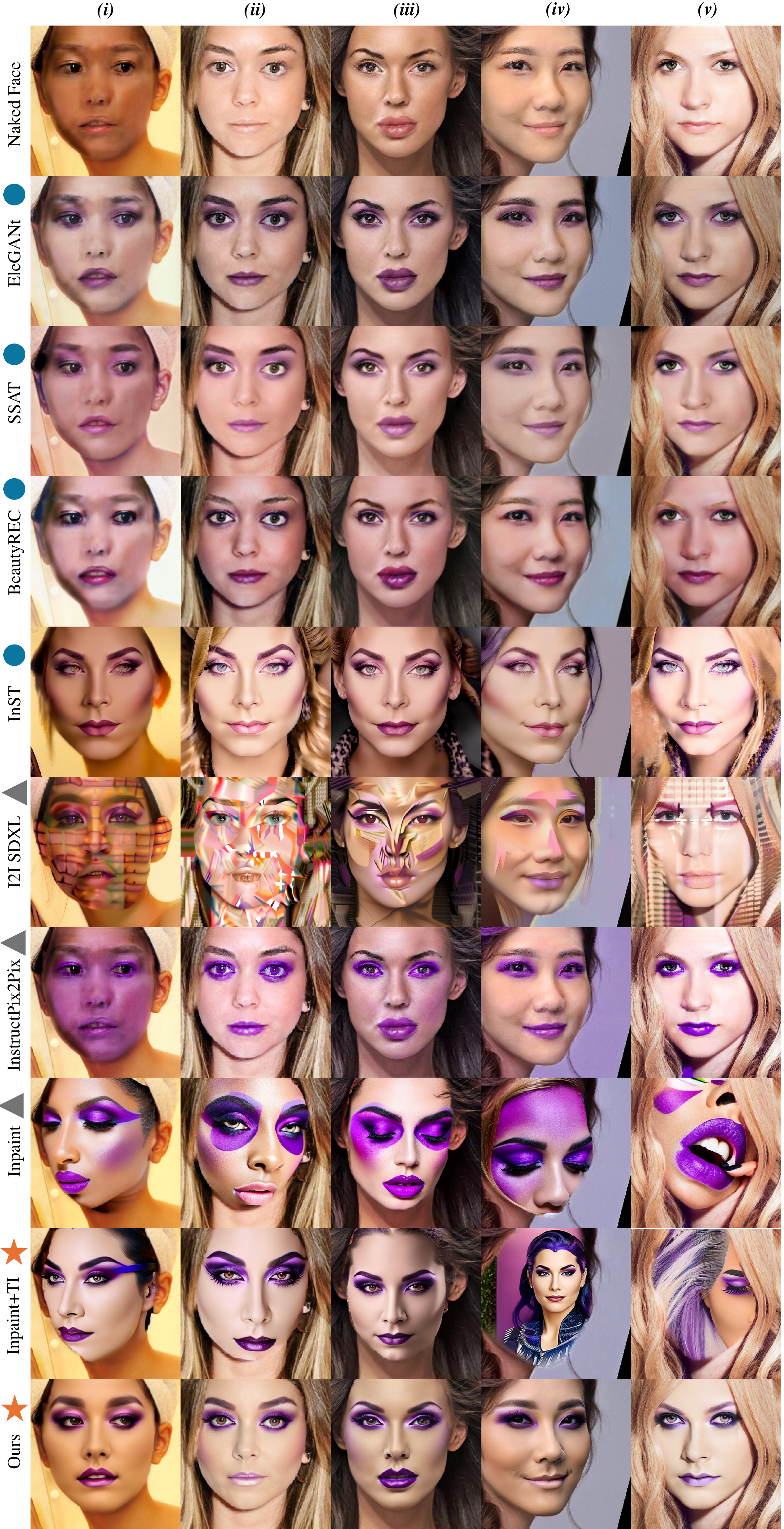}
	\caption{Comparative qualitative results for Style 1 (d), showcasing the performance of our method against other baseline approaches. \textit{Note:} The \textit{blue circle} signifies that the method utilized the specific image referenced by the blue circle in Fig. \ref{fig:style1_bank}. Methods annotated with an \textit{orange star} have drawn inspiration from the images indicated by the orange star in Fig. \ref{fig:style1_bank}. In contrast, a \textit{gray triangle} denotes methods that rely solely on a textual description, specifically the prompt ``A photo of a woman with purple makeup on face,'' to guide the makeup generation process.}
	\label{fig:style1d}
\end{figure}

\begin{figure}[ht]
	\centering 
	\includegraphics[width=0.5\textwidth]{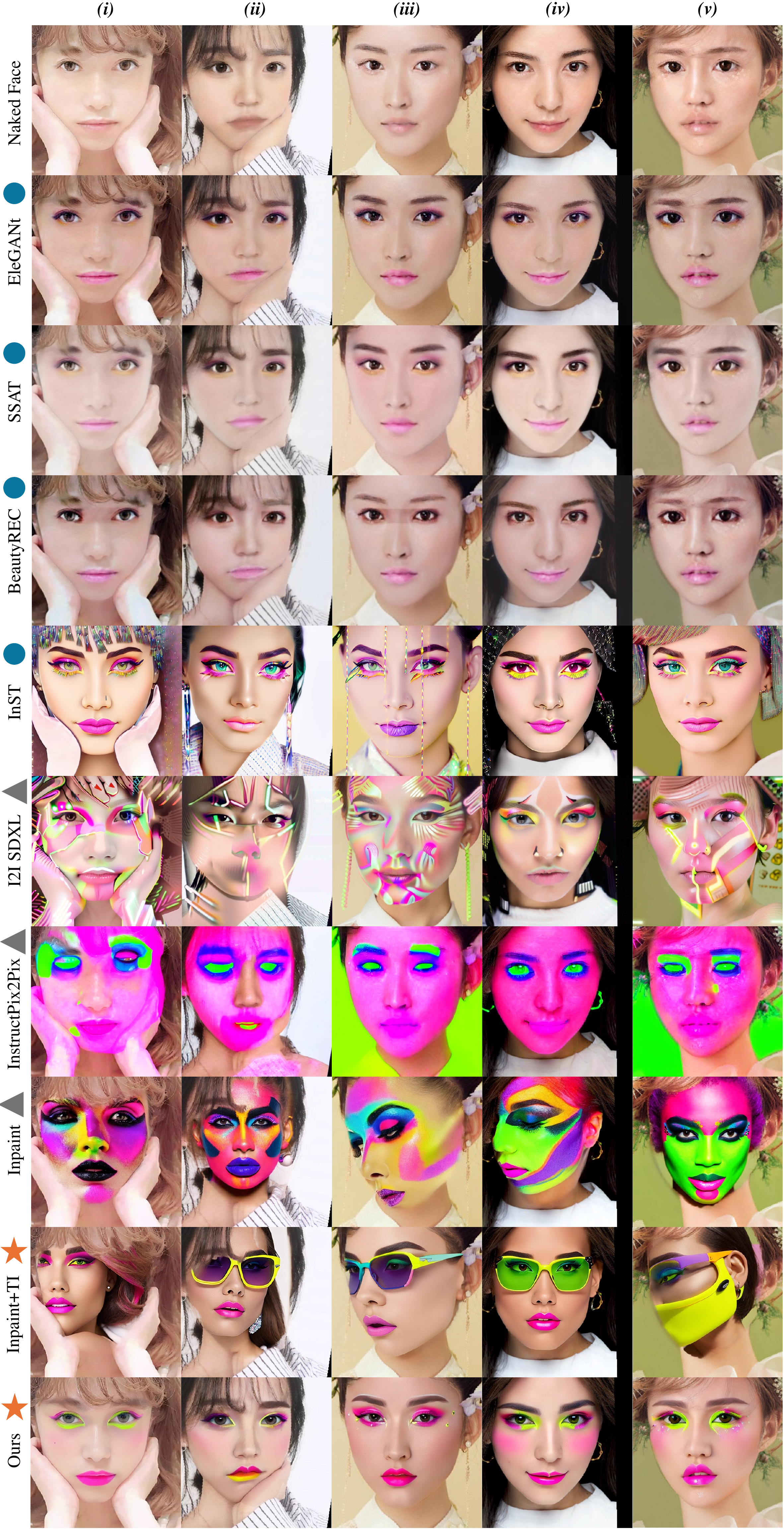}
	\caption{Comparative qualitative results for Style 1 (e), showcasing the performance of our method against other baseline approaches. \textit{Note:} The \textit{blue circle} signifies that the method utilized the specific image referenced by the blue circle in Fig. \ref{fig:style1_bank}. Methods annotated with an \textit{orange star} have drawn inspiration from the images indicated by the orange star in Fig. \ref{fig:style1_bank}. In contrast, a \textit{gray triangle} denotes methods that rely solely on a textual description, specifically the prompt ``A photo of a woman with neon makeup on face,'' to guide the makeup generation process.}
	\label{fig:style1e}
\end{figure}

\begin{figure}[ht]
	\centering 
	\includegraphics[width=0.5\textwidth]{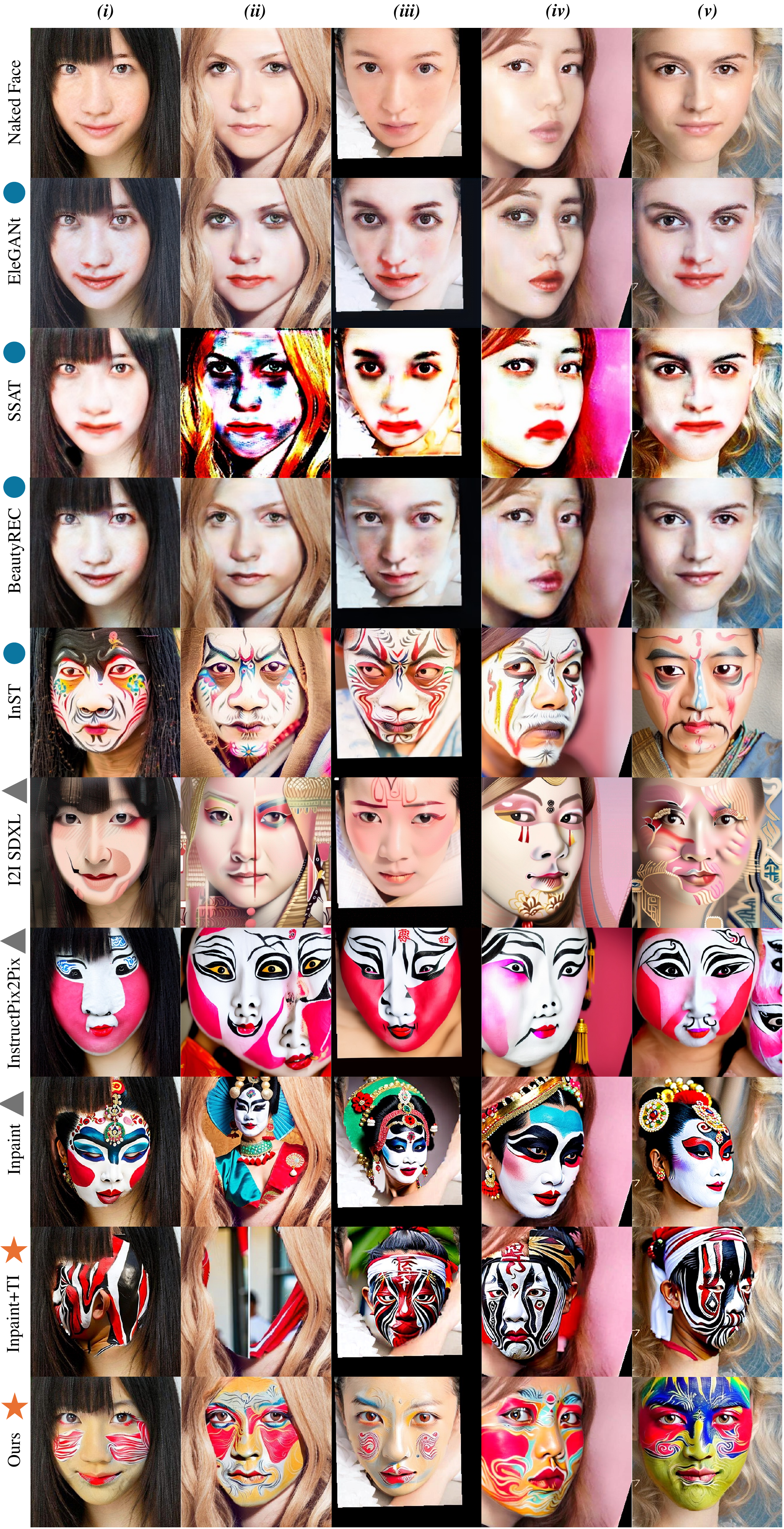}
	\caption{Comparative qualitative results for Style 1 (f), showcasing the performance of our method against other baseline approaches. \textit{Note:} The \textit{blue circle} signifies that the method utilized the specific image referenced by the blue circle in Fig. \ref{fig:style1_bank}. Methods annotated with an \textit{orange star} have drawn inspiration from the images indicated by the orange star in Fig. \ref{fig:style1_bank}. In contrast, a \textit{gray triangle} denotes methods that rely solely on a textual description, specifically the prompt ``A photo of a woman with peking opera makeup on face,'' to guide the makeup generation process.}
	\label{fig:style1f}
\end{figure}

\begin{figure}[ht]
	\centering 
	\includegraphics[width=0.5\textwidth]{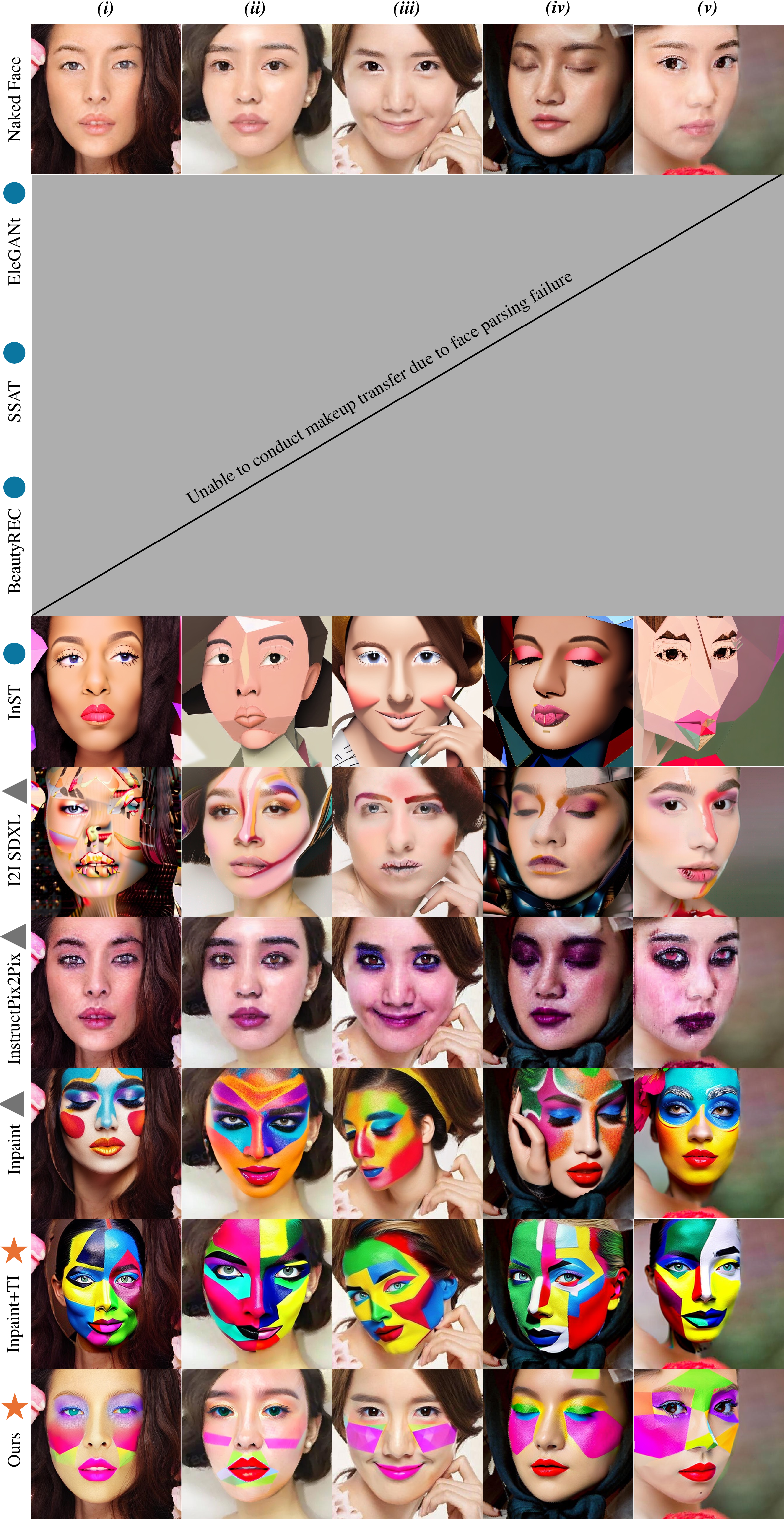}
	\caption{Comparative qualitative results for Style 2 (a), showcasing the performance of our method against other baseline approaches. \textit{Note:} The \textit{blue circle} signifies that the method utilized the specific image referenced by the blue circle in Fig. \ref{fig:style1_bank}. Methods annotated with an \textit{orange star} have drawn inspiration from the images indicated by the orange star in Fig. \ref{fig:style1_bank}. In contrast, a \textit{gray triangle} denotes methods that rely solely on a textual description, specifically the prompt ``A photo of a woman with art makeup on face,'' to guide the makeup generation process.}
	\label{fig:style2a}
\end{figure}

\begin{figure}[ht]
	\centering 
	\includegraphics[width=0.5\textwidth]{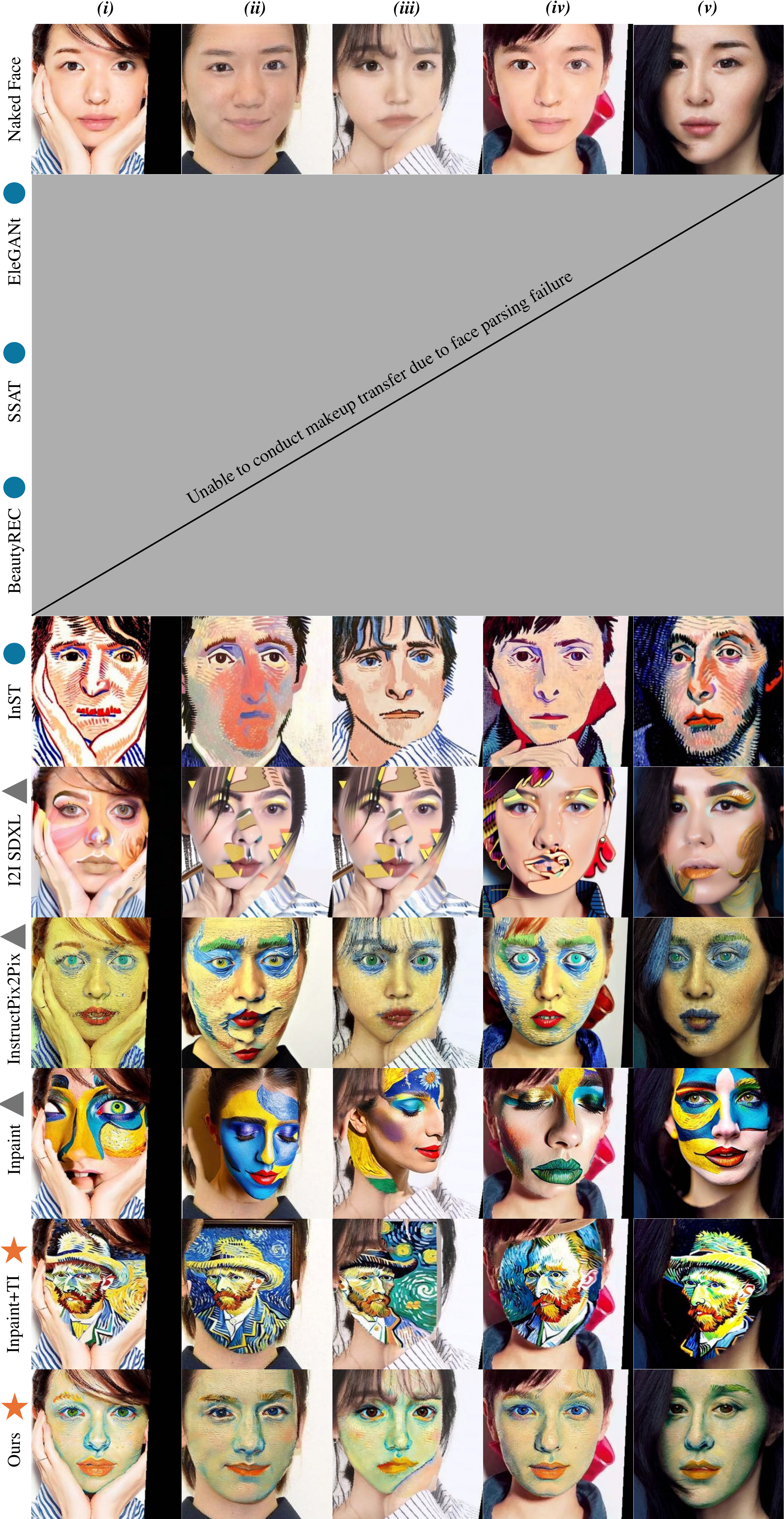}
	\caption{Comparative qualitative results for Style 2 (b), showcasing the performance of our method against other baseline approaches. \textit{Note:} The \textit{blue circle} signifies that the method utilized the specific image referenced by the blue circle in Fig. \ref{fig:style1_bank}. Methods annotated with an \textit{orange star} have drawn inspiration from the images indicated by the orange star in Fig. \ref{fig:style1_bank}. In contrast, a \textit{gray triangle} denotes methods that rely solely on a textual description, specifically the prompt ``A photo of a woman with Van Gogh makeup on face,'' to guide the makeup generation process.}
	\label{fig:style2b}
\end{figure}

\begin{figure}[ht]
	\centering 
	\includegraphics[width=0.5\textwidth]{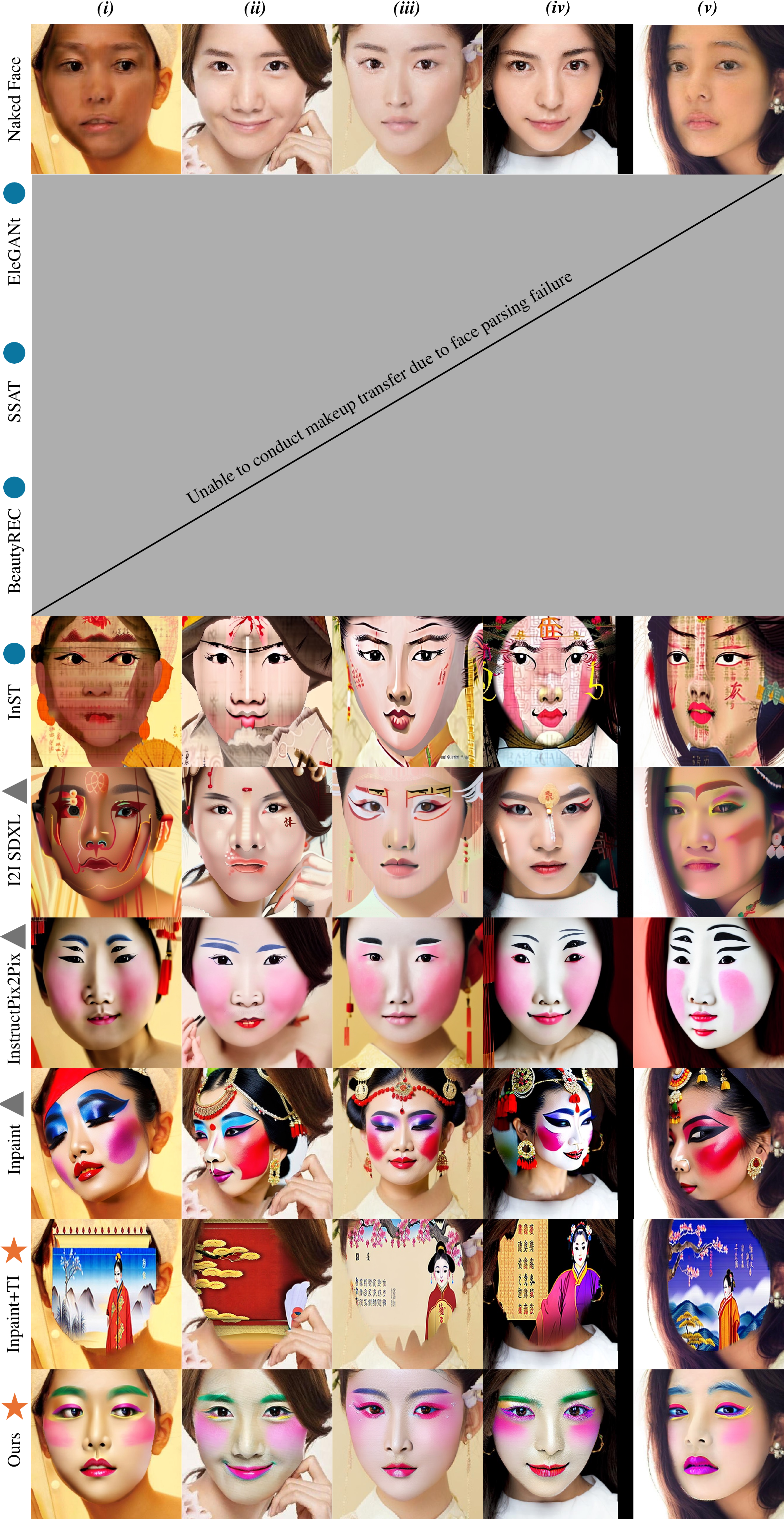}
	\caption{Comparative qualitative results for Style 2 (c), showcasing the performance of our method against other baseline approaches. \textit{Note:} The \textit{blue circle} signifies that the method utilized the specific image referenced by the blue circle in Fig. \ref{fig:style1_bank}. Methods annotated with an \textit{orange star} have drawn inspiration from the images indicated by the orange star in Fig. \ref{fig:style1_bank}. In contrast, a \textit{gray triangle} denotes methods that rely solely on a textual description, specifically the prompt ``A photo of a woman with traditional Chinese makeup on face,'' to guide the makeup generation process.}
	\label{fig:style2c}
\end{figure}

\begin{figure}[ht]
	\centering 
	\includegraphics[width=0.5\textwidth]{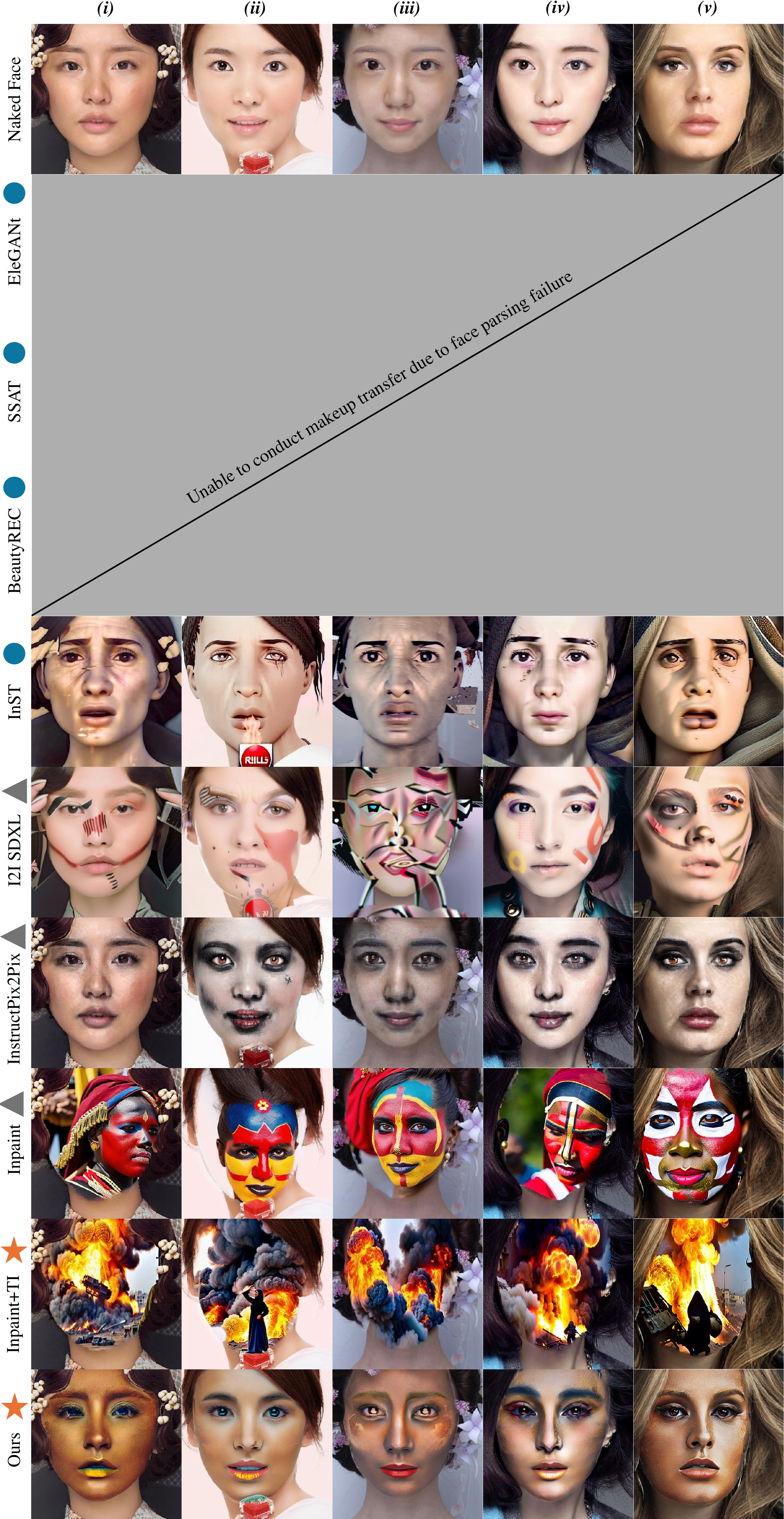}
	\caption{Comparative qualitative results for Style 2 (d), showcasing the performance of our method against other baseline approaches. \textit{Note:} The \textit{blue circle} signifies that the method utilized the specific image referenced by the blue circle in Fig. \ref{fig:style1_bank}. Methods annotated with an \textit{orange star} have drawn inspiration from the images indicated by the orange star in Fig. \ref{fig:style1_bank}. In contrast, a \textit{gray triangle} denotes methods that rely solely on a textual description, specifically the prompt ``A photo of a woman with war makeup on face,'' to guide the makeup generation process.}
	\label{fig:style2d}
\end{figure}

\begin{figure}[ht]
	\centering 
	\includegraphics[width=0.5\textwidth]{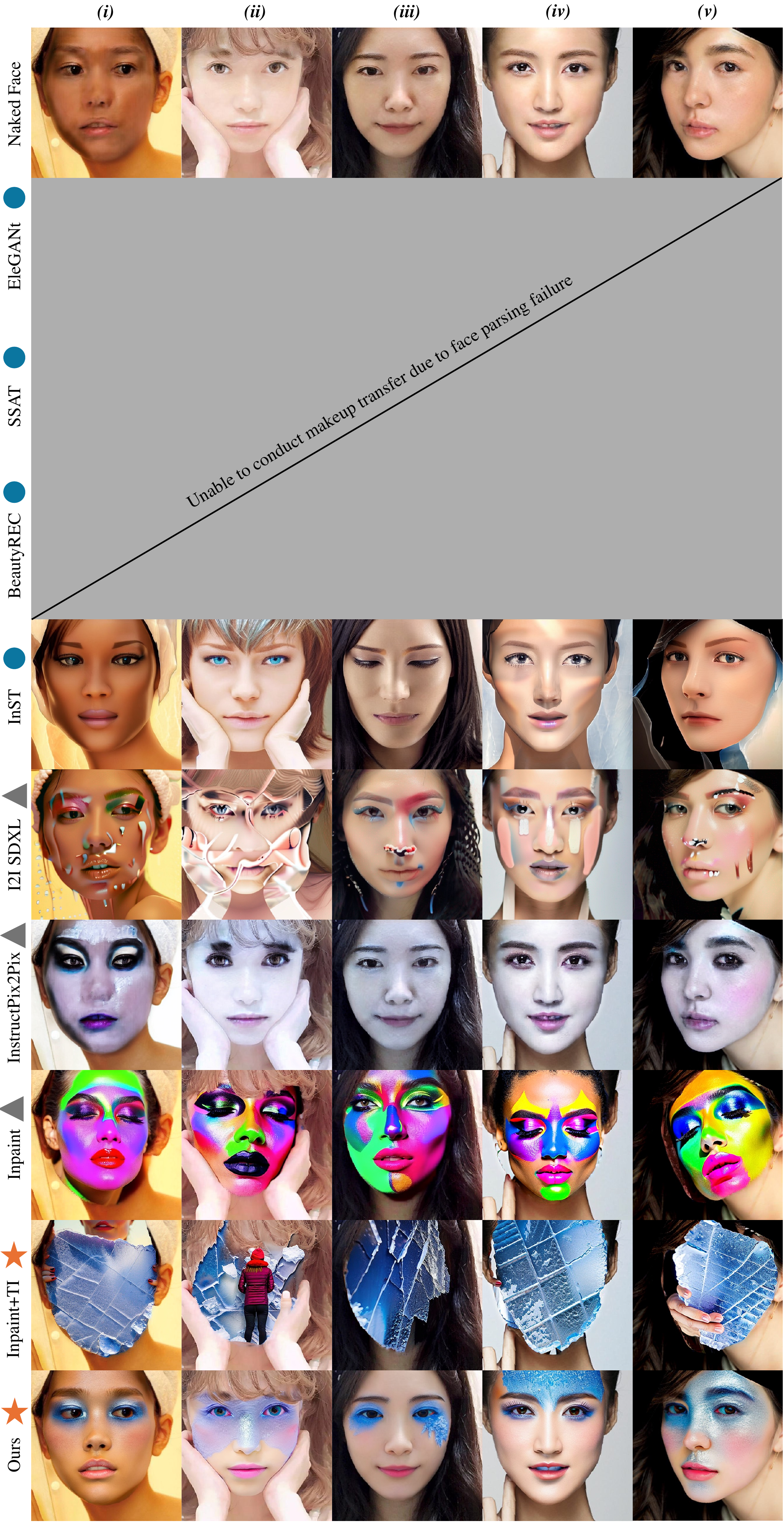}
	\caption{Comparative qualitative results for Style 2 (e), showcasing the performance of our method against other baseline approaches. \textit{Note:} The \textit{blue circle} signifies that the method utilized the specific image referenced by the blue circle in Fig. \ref{fig:style1_bank}. Methods annotated with an \textit{orange star} have drawn inspiration from the images indicated by the orange star in Fig. \ref{fig:style1_bank}. In contrast, a \textit{gray triangle} denotes methods that rely solely on a textual description, specifically the prompt ``A photo of a woman with ice makeup on face,'' to guide the makeup generation process.}
	\label{fig:style2e}
\end{figure}

\begin{figure}[ht]
	\centering 
	\includegraphics[width=0.5\textwidth]{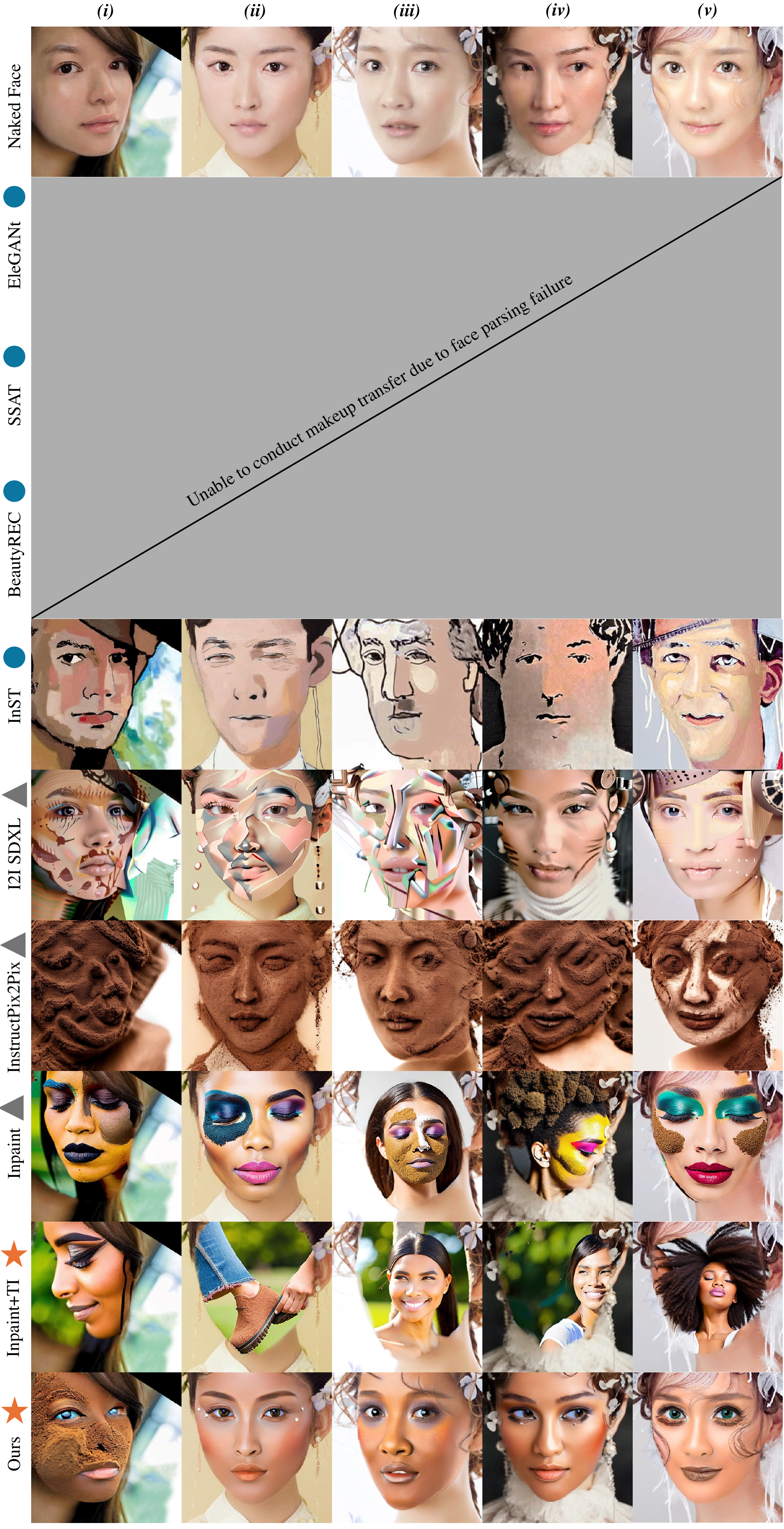}
	\caption{Comparative qualitative results for Style 2 (f), showcasing the performance of our method against other baseline approaches. \textit{Note:} The \textit{blue circle} signifies that the method utilized the specific image referenced by the blue circle in Fig. \ref{fig:style1_bank}. Methods annotated with an \textit{orange star} have drawn inspiration from the images indicated by the orange star in Fig. \ref{fig:style1_bank}. In contrast, a \textit{gray triangle} denotes methods that rely solely on a textual description, specifically the prompt ``A photo of a woman with soil makeup on face,'' to guide the makeup generation process.}
	\label{fig:style2f}
\end{figure}

\begin{figure}[ht]
	\centering 
	\includegraphics[width=0.5\textwidth]{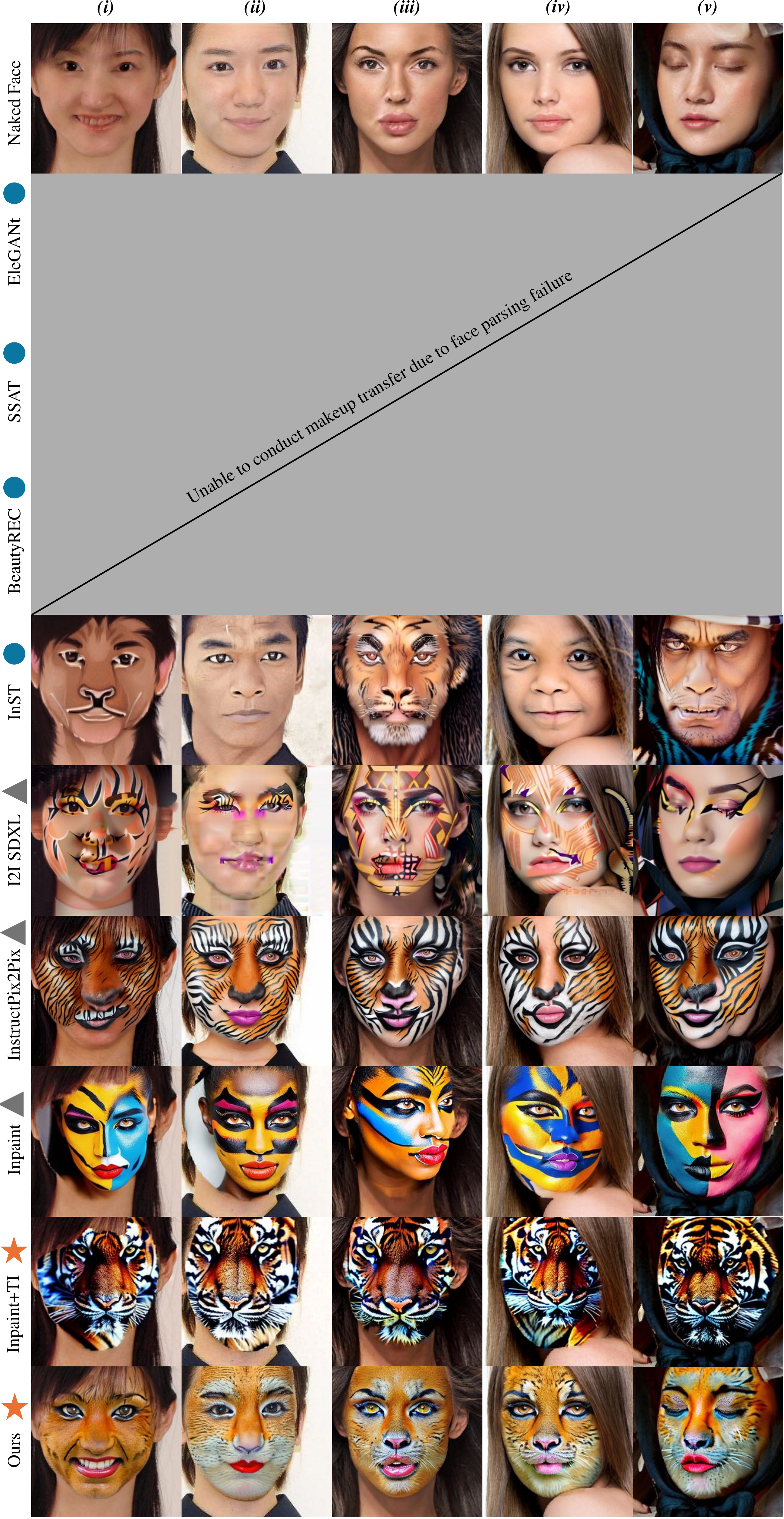}
	\caption{Comparative qualitative results for Style 2 (g), showcasing the performance of our method against other baseline approaches. \textit{Note:} The \textit{blue circle} signifies that the method utilized the specific image referenced by the blue circle in Fig. \ref{fig:style1_bank}. Methods annotated with an \textit{orange star} have drawn inspiration from the images indicated by the orange star in Fig. \ref{fig:style1_bank}. In contrast, a \textit{gray triangle} denotes methods that rely solely on a textual description, specifically the prompt ``A photo of a woman with tiger makeup on face,'' to guide the makeup generation process.}
	\label{fig:style2g}
\end{figure}

\begin{figure}[ht]
	\centering 
	\includegraphics[width=0.5\textwidth]{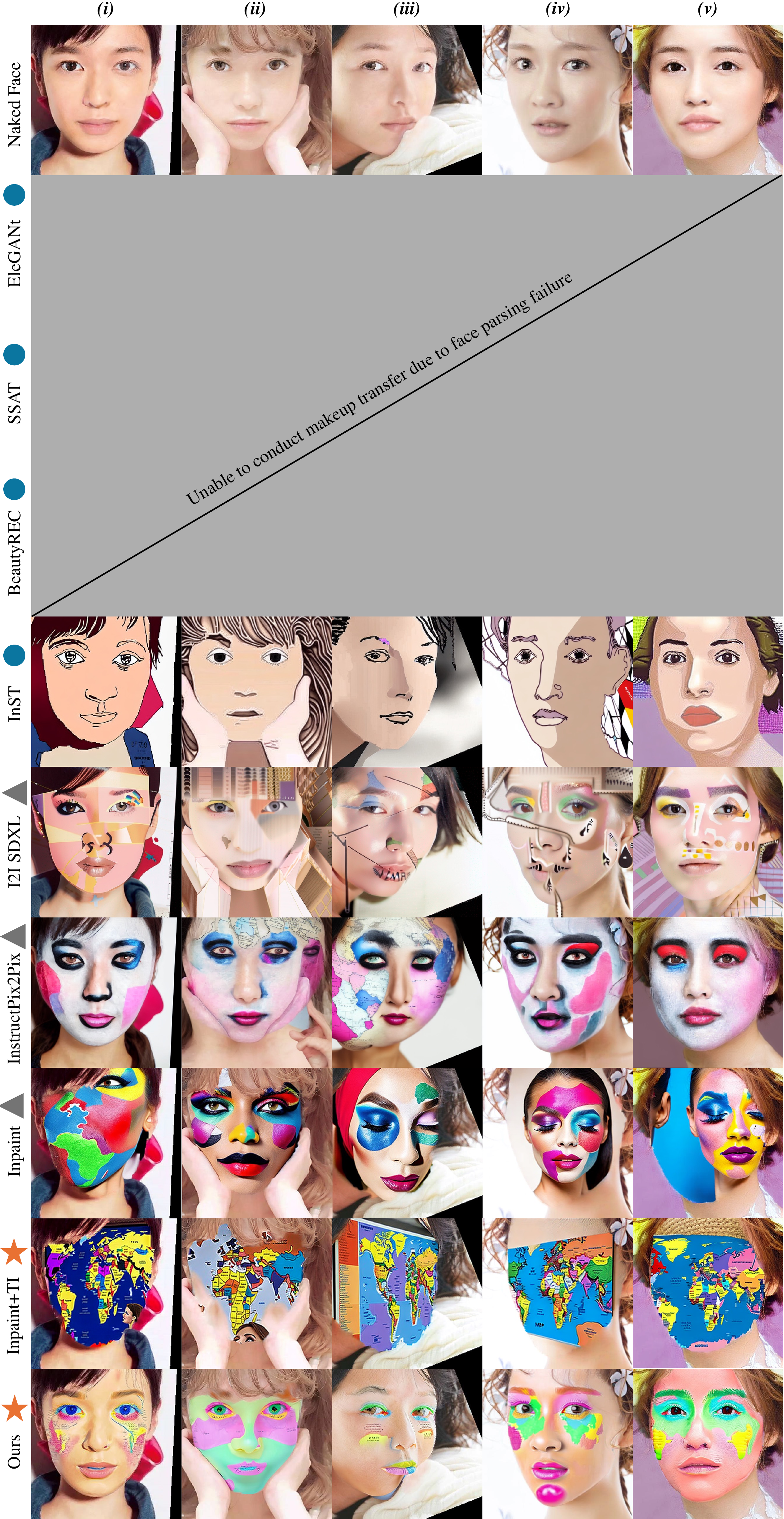}
	\caption{Comparative qualitative results for Style 2 (h), showcasing the performance of our method against other baseline approaches. \textit{Note:} The \textit{blue circle} signifies that the method utilized the specific image referenced by the blue circle in Fig. \ref{fig:style1_bank}. Methods annotated with an \textit{orange star} have drawn inspiration from the images indicated by the orange star in Fig. \ref{fig:style1_bank}. In contrast, a \textit{gray triangle} denotes methods that rely solely on a textual description, specifically the prompt ``A photo of a woman with world map makeup on face,'' to guide the makeup generation process.}
	\label{fig:style2h}
\end{figure}

\begin{figure}[ht]
	\centering 
	\includegraphics[width=0.5\textwidth]{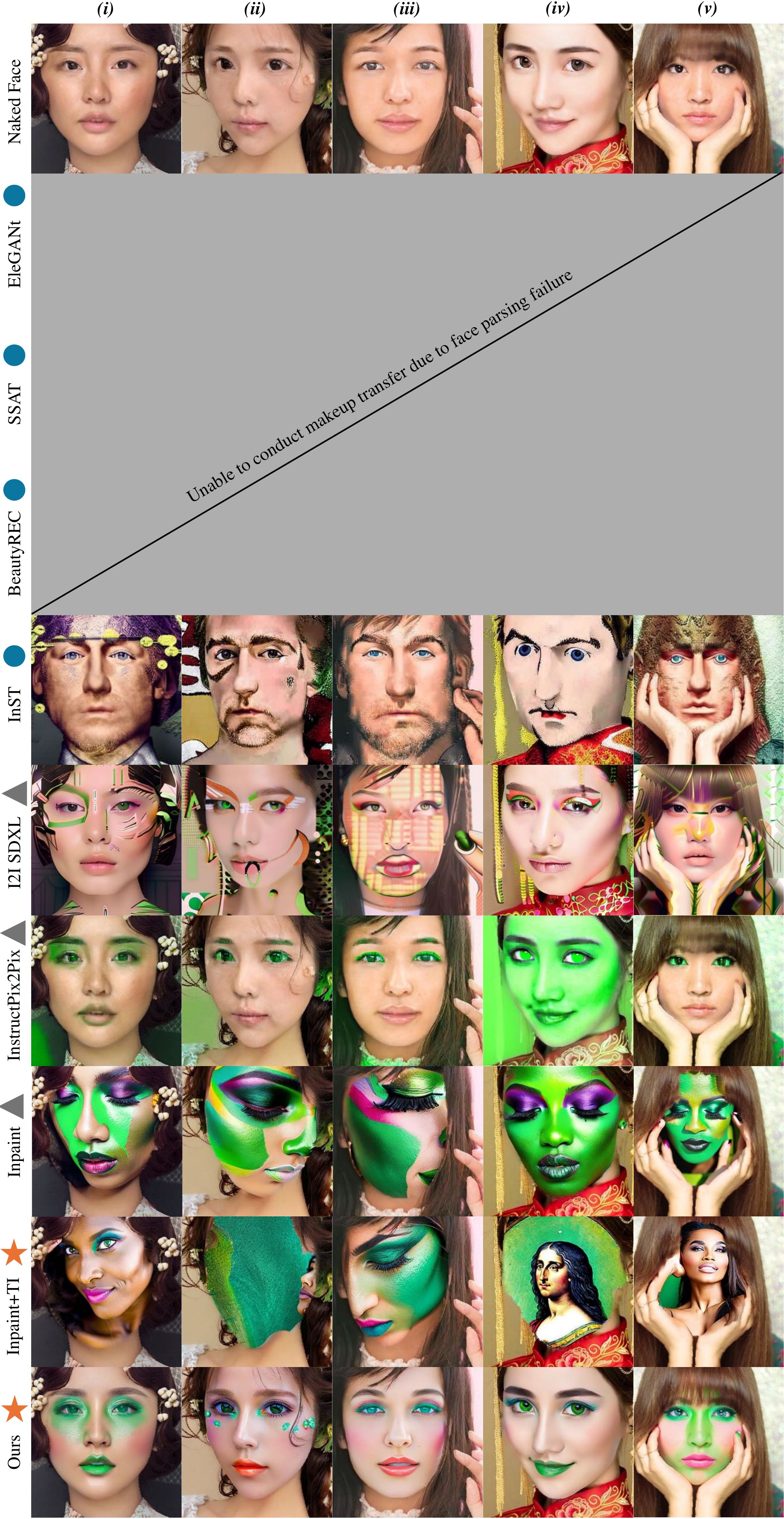}
	\caption{Comparative qualitative results for Style 2 (i), showcasing the performance of our method against other baseline approaches. \textit{Note:} The \textit{blue circle} signifies that the method utilized the specific image referenced by the blue circle in Fig. \ref{fig:style1_bank}. Methods annotated with an \textit{orange star} have drawn inspiration from the images indicated by the orange star in Fig. \ref{fig:style1_bank}. In contrast, a \textit{gray triangle} denotes methods that rely solely on a textual description, specifically the prompt ``A photo of a woman with green makeup on face,'' to guide the makeup generation process.}
	\label{fig:style2i}
\end{figure}

\begin{figure}[ht]
	\centering 
	\includegraphics[width=0.5\textwidth]{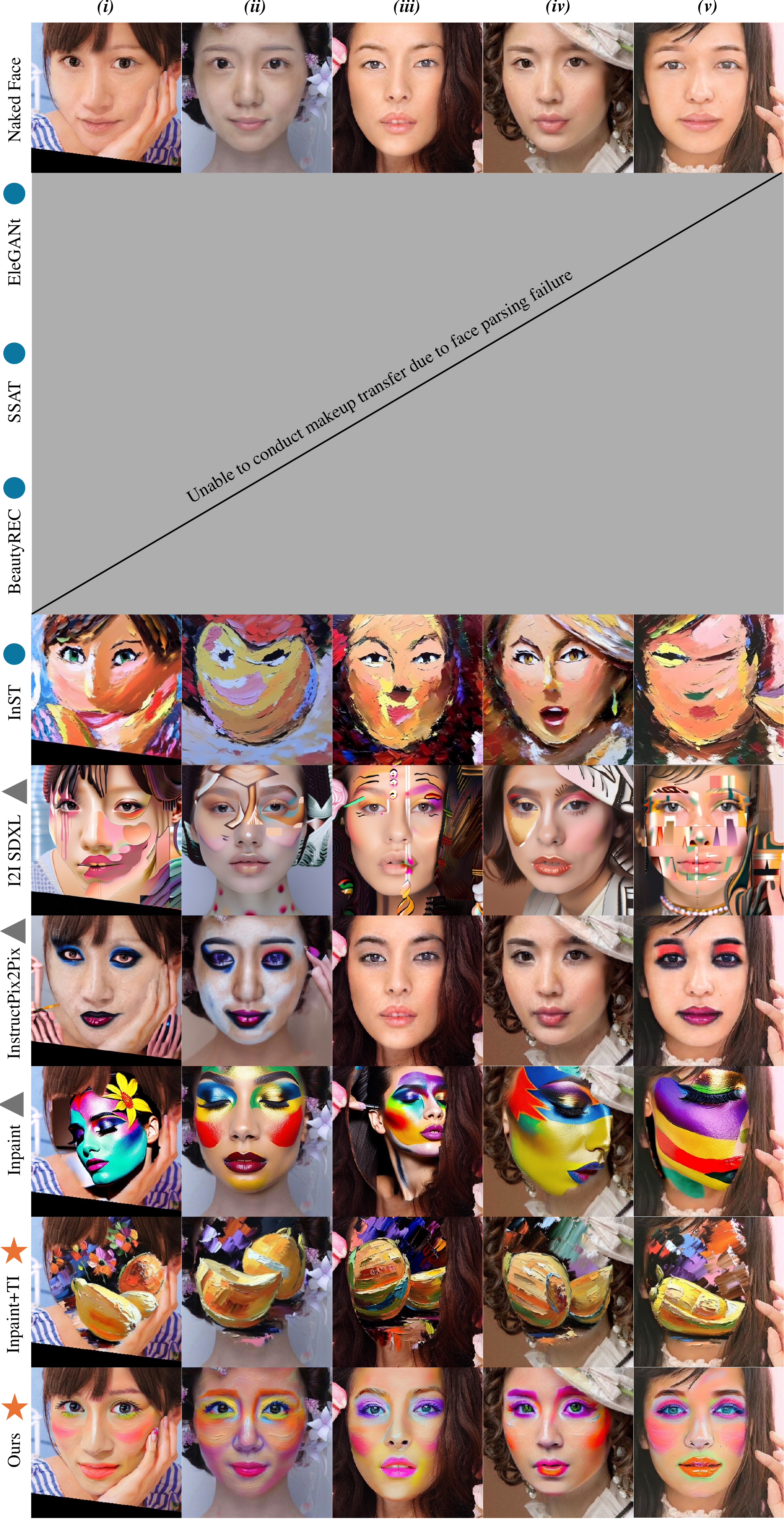}
	\caption{Comparative qualitative results for Style 2 (j), showcasing the performance of our method against other baseline approaches. \textit{Note:} The \textit{blue circle} signifies that the method utilized the specific image referenced by the blue circle in Fig. \ref{fig:style1_bank}. Methods annotated with an \textit{orange star} have drawn inspiration from the images indicated by the orange star in Fig. \ref{fig:style1_bank}. In contrast, a \textit{gray triangle} denotes methods that rely solely on a textual description, specifically the prompt ``A photo of a woman with oil painting makeup on face,'' to guide the makeup generation process.}
	\label{fig:style2j}
\end{figure}

\begin{figure}[ht]
	\centering 
	\includegraphics[width=0.5\textwidth]{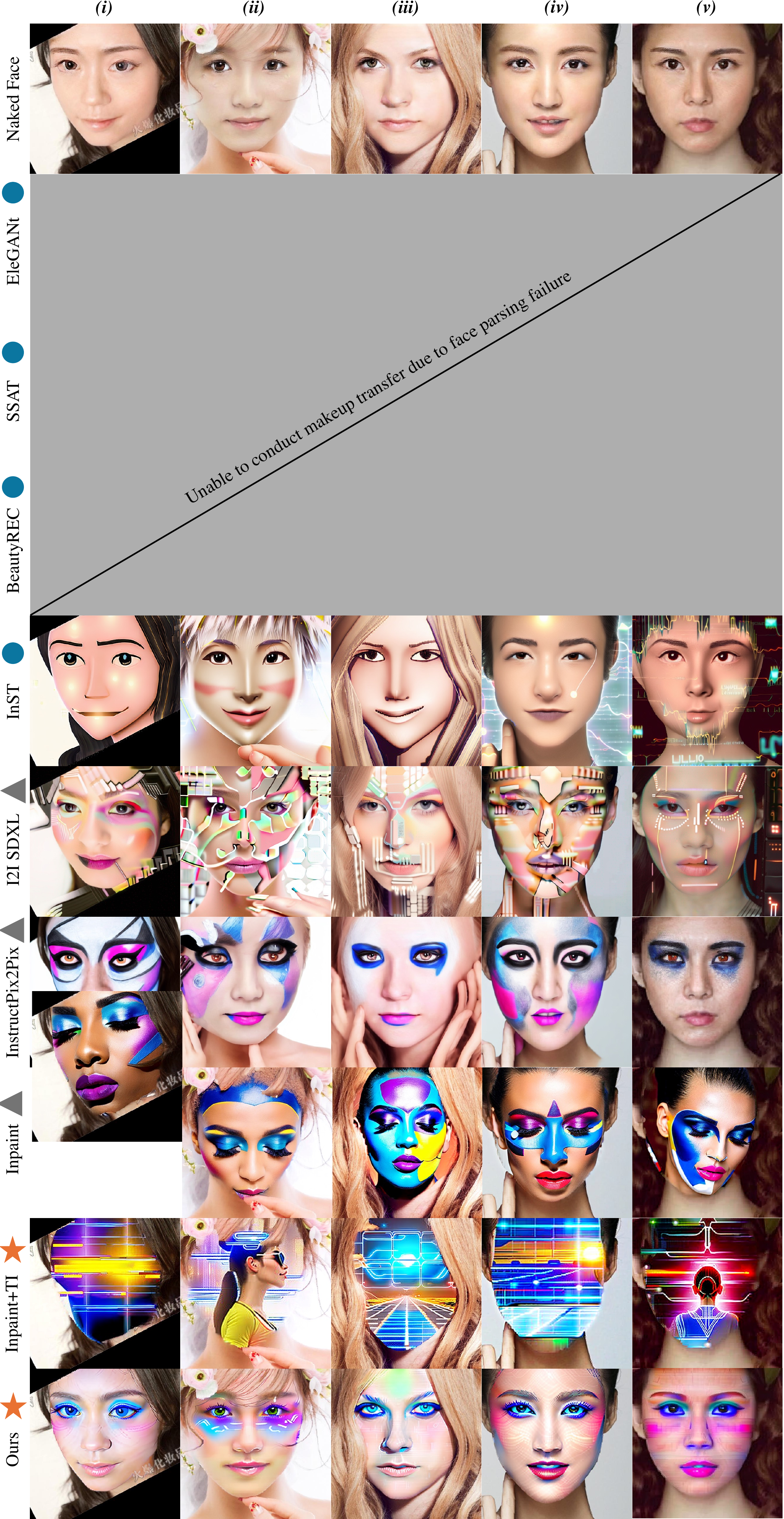}
	\caption{Comparative qualitative results for Style 2 (k), showcasing the performance of our method against other baseline approaches. \textit{Note:} The \textit{blue circle} signifies that the method utilized the specific image referenced by the blue circle in Fig. \ref{fig:style1_bank}. Methods annotated with an \textit{orange star} have drawn inspiration from the images indicated by the orange star in Fig. \ref{fig:style1_bank}. In contrast, a \textit{gray triangle} denotes methods that rely solely on a textual description, specifically the prompt ``A photo of a woman with technology makeup on face,'' to guide the makeup generation process.}
	\label{fig:style2k}
\end{figure}

\begin{figure}[ht]
	\centering 
	\includegraphics[width=0.5\textwidth]{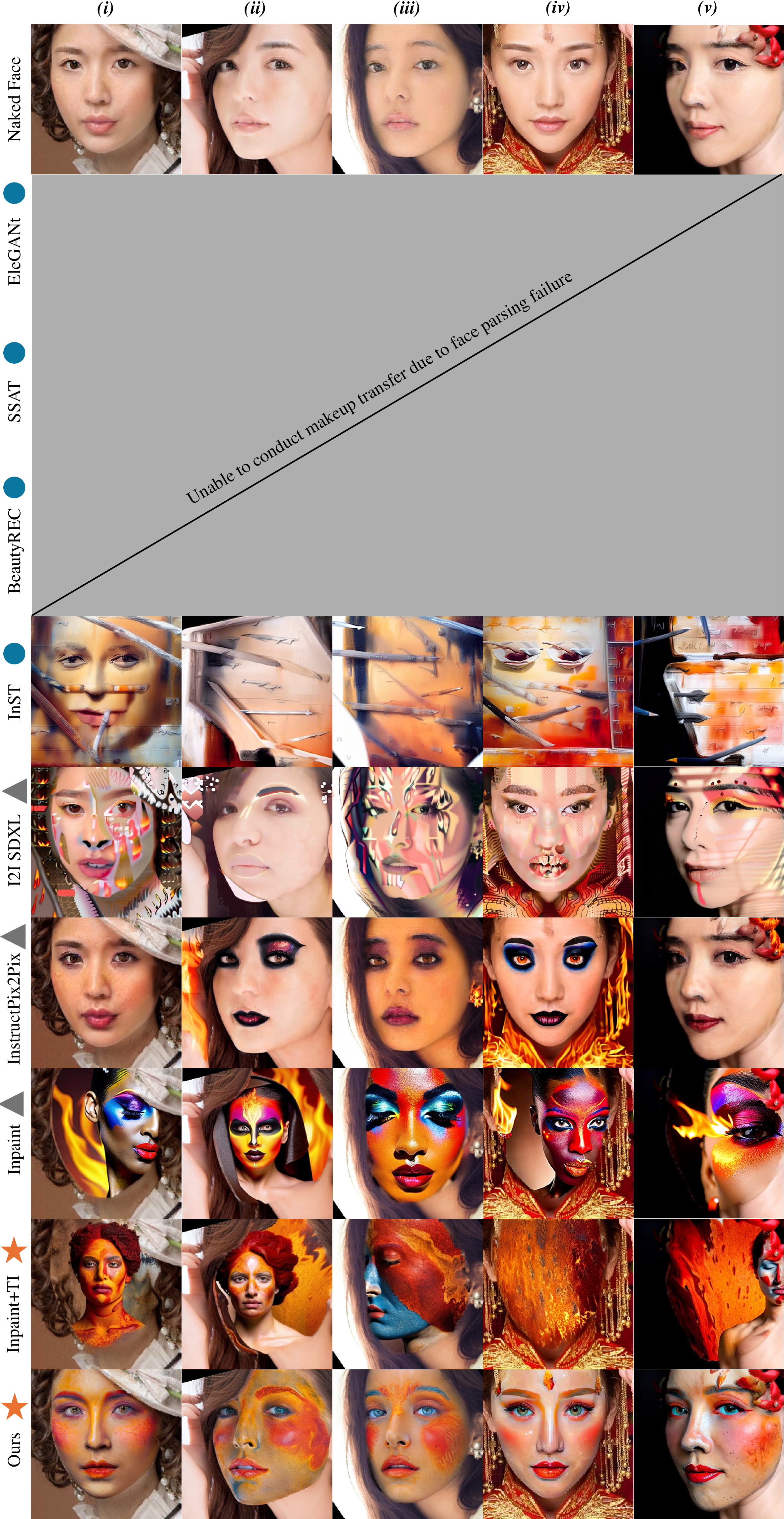}
	\caption{Comparative qualitative results for Style 2 (l), showcasing the performance of our method against other baseline approaches. \textit{Note:} The \textit{blue circle} signifies that the method utilized the specific image referenced by the blue circle in Fig. \ref{fig:style1_bank}. Methods annotated with an \textit{orange star} have drawn inspiration from the images indicated by the orange star in Fig. \ref{fig:style1_bank}. In contrast, a \textit{gray triangle} denotes methods that rely solely on a textual description, specifically the prompt ``A photo of a woman with fire makeup on face,'' to guide the makeup generation process.}
	\label{fig:style2l}
\end{figure}

This section presents further qualitative results produced by $Gorgeous$ for Style 1 and Style 2 ideas, as illustrated in Figures \ref{fig:style1a} to \ref{fig:style1f} and Figures \ref{fig:style2a} to \ref{fig:style2l}, respectively. These results are benchmarked against other baseline methods in the field.
\begin{figure*}[th]
	\centering 
	\includegraphics[width=0.85\linewidth]{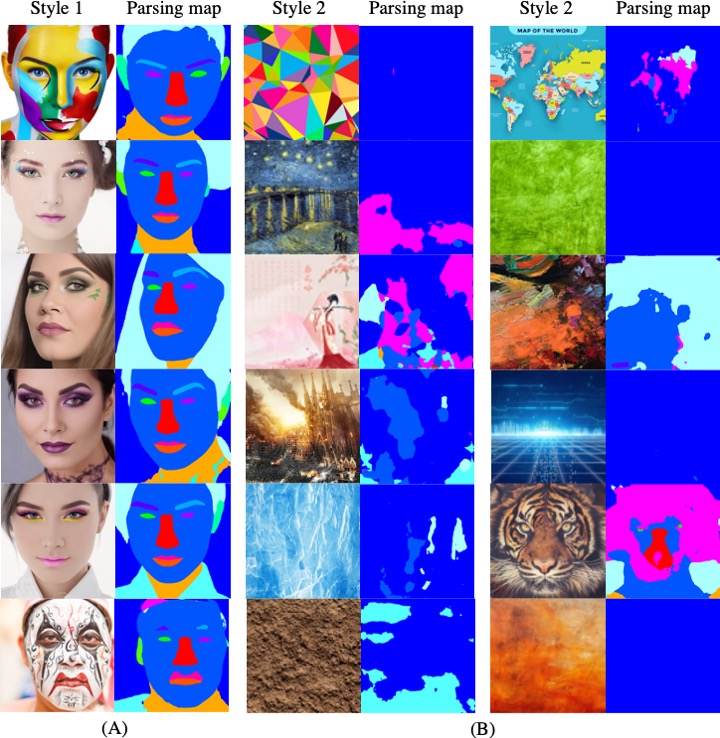}
	\caption{Comparative demonstration of face parsing performance on facial versus non-facial reference images for makeup transfer. (A) showcases the successful parsing of facial features in reference images containing faces, facilitating effective makeup transfer. Conversely, (B) illustrates the failure of face parsing when applied to non-facial reference images; the lack of detectable facial features leads to inaccurate parsing maps, as indicated by the absence of correctly identified face regions. This parsing failure precludes the direct application of makeup transfer techniques to such images, necessitating alternative methods like $Gorgeous$ that can interpret and utilize non-facial stylistic elements for makeup generation without reliance on facial feature detection.}
	\label{fig:parsing}
\end{figure*}

Key observations from our evaluations include:
\begin{enumerate}[label=(\roman*)]
    \item \textbf{Unique Character Creations:} $Gorgeous$ consistently generates unique character facial makeups that stand apart from traditional makeup transfer methods such as EleGANt \cite{yang2022elegant}, SSAT \cite{sun2022ssat}, and BeautyREC \cite{yan2023beautyrec}.
    \item \textbf{Fidelity and Accuracy:} Traditional makeup transfer methods often struggle to replicate the fidelity of makeup styles from reference images, particularly when styles are exaggerated or highly stylized (e.g., Style 1(a) and Style 1(f)). This underscores a significant potential for improvement in conventional approaches. Therefore, results obtained from traditional makeup transfer methods can only be a light reference in our case. 
    \item \textbf{Adaptation to Non-Facial Styles:} Our results also highlight the successful adaptation of non-facial style inspirations into facial makeup, as shown in Figures \ref{fig:style2a} to \ref{fig:style2l}. Unlike existing methods that rely heavily on accurate face parsing—which often fails with non-traditional styles—$Gorgeous$ effectively bridges this gap. The failure of face parsing is detailed in Figure \ref{fig:parsing}, showcasing the limitations of traditional methods in handling non-conventional makeup styles.
    \item \textbf{Comparison with Other Methods:} Further comparisons with diffusion-based and text-guided image generation models (e.g., SDXL \cite{Podell2023SDXLIL}, InstructPix2Pix \cite{Brooks2022InstructPix2PixLT}, and Stable Diffusion \cite{Rombach_2022_CVPR}) reveal that while these methods are innovative, they typically struggle to preserve the original facial identity during the style translation process.

$Gorgeous$ not only transforms stylistic elements from any image into applicable makeup formats but also demonstrates superior flexibility and creativity over traditional makeup transfer, style transfer and other relevant generation methods. It excels in producing distinctive character looks that both maintain the identity of the original face and the integrity of non-facial areas. This capability sets a new standard in the field of character facial makeup generation, showing our method's advanced ability to handle diverse image types and complex makeup challenges.
\end{enumerate}

\clearpage

\clearpage

\clearpage
\onecolumn
\subsection{Extended Quantitative Evaluations for Style 1 and 2}
\begin{table*}[ht]
\centering
\caption{Expanded quantitative evaluation of our method with relevant competitive methods. **The absence of makeup transfer scores for Style 2(a-l) is due to the failure of existing makeup transfer methods to transfer the makeups, given a reference image without a human face. Both CSD and DreamSIM are measured in cosine similarity. Bold value indicates the best score in the column.}
\begin{tabular}{lcccccc}
\toprule
\textbf{Methods} & \multicolumn{3}{c}{\textbf{Average of Style 1(a-f)}} & \multicolumn{3}{c}{\textbf{Average of Style 2(a-l)}} \\
& CSD $\uparrow$ & DreamSIM $\uparrow$ & FID $\downarrow$ & CSD $\uparrow$ & DreamSIM $\uparrow$ & FID $\downarrow$  \\
\midrule
\multicolumn{7}{l}{\textbf{(Makeup transfer)}}\\
EleGANt \cite{yang2022elegant}  & 0.41 & 0.51 & 41.96 & - & - & - \\
SSAT \cite{sun2022ssat}  & 0.41 & 0.53 & 57.40 & - & - & -\\
BeautyREC \cite{yan2023beautyrec} & 0.34 & 0.46 & {\textbf{39.63}} & - & - & -\\
\midrule
\multicolumn{7}{l}{\textbf{(Style transfer)}}\\
InST \cite{zhang2023inversion} & 0.36 & {\textbf{0.66}} & 136.65 & 0.24 & 0.25 & 193.02 \\
\midrule
\multicolumn{7}{l}{\textbf{(Image-to-image translation/generation)}}\\
I2I SDXL \cite{Podell2023SDXLIL} & 0.44 & 0.53 & 145.78 & 0.13 & 0.19 & 147.69 \\
InstructPix2Pix \cite{Brooks2022InstructPix2PixLT} & 0.52 & 0.54 & 146.54 & 0.21 & 0.20 & 115.06 \\
Inpainting \cite{Rombach_2022_CVPR} & 0.55 & 0.54 & 166.07 & 0.16 & 0.21 & 168.06 \\
Inpainting+TI \cite{Rombach_2022_CVPR,Gal2022AnII} & {\textbf{0.58}} & 0.63 & 154.29 & {\textbf{0.31}} & {\textbf{0.41}} & 231.57 \\
\midrule
Ours (Gorgeous) & {\textbf{0.58}} & 0.60 & 80.87 & 0.19 & 0.25 & {\textbf{84.84}} \\
\bottomrule
\end{tabular}
\label{tab:quan_supp}
\end{table*}
The quantitative results presented in Tab. \ref{tab:quan_supp} further affirm the robust performance of $Gorgeous$ across an expanded set of themed examples, as discussed in Section 6. These evaluations demonstrate the consistency of our method in handling diverse makeup styles.

\begin{enumerate}[label=(\roman*)]
    \item \textbf{Superiority in FID Scores:} $Gorgeous$ consistently achieves the lowest FID \cite{heusel2017gans} scores among competing image generation models, recording 80.87 for Style 1 and 84.84 for Style 2. This indicates superior ability of $Gorgeous$ to translate inspirational images into makeup formats that closely align with the nuances of the BeautyFace dataset \cite{yan2023beautyrec}. The model also shows robust performance in CSD \cite{Somepalli2024MeasuringSS} and DreamSIM \cite{Fu2023DreamSimLN} metrics, further confirming its effectiveness in capturing and replicating style relevance.
    \item \textbf{Comparison with Makeup Transfer Methods:} Traditional makeup transfer methods like BeautyREC \cite{yan2023beautyrec} (FID 39.63), EleGANt \cite{yang2022elegant} (FID 41.96), and SSAT \cite{sun2022ssat} (FID 57.40) may exhibit lower FID scores in Style 1, but they lack the ability to generate diverse and unique makeup styles. Instead, they primarily replicate existing looks, showing a lack of innovation. For Style 2, which includes non-facial references, these methods were inapplicable due to their reliance on facial parsing, highlighting a critical limitation in versatility.
    \item \textbf{Style Transfer Method Performance:} The Style Transfer method, InST \cite{zhang2023inversion}, while scoring high in DreamSIM (0.66 for Style 1 and 0.25 for Style 2) and decently in CSD (0.36 for Style 1 and 0.24 for Style 2), significantly underperforms in FID (136.65 for Style 1 and 193.02 for Style 2), suggesting a divergence from accurate makeup application in terms of adhering to a standard makeup format.
    \item \textbf{Image-to-Image Translation and Generation Challenges:} Inpainting \cite{Rombach_2022_CVPR} combined with Textual Inversion \cite{Gal2022AnII} demonstrates commendable scores in style capture—CSD (0.58 for Style 1, 0.31 for Style 2) and DreamSIM (0.63 for Style 1, 0.41 for Style 2). However, it underachieves in maintaining low FID scores (154.29 for Style 1, 231.57 for Style 2), indicating that while it captures the style from references well, it struggles to preserve the integrity of the facial identity and fails to render the style as a practical, wearable makeup.
\end{enumerate}

In conclusion, $Gorgeous$ excels across all key performance metrics, particularly in its ability to create authentic character looks that faithfully match the intended makeup formats from the BeautyFace dataset. This underscores its superior capability to adapt to and creatively transform a broad range of image inspirations into practical makeup applications.

\clearpage
\subsection{Demonstration of $Gorgeous$ on Wild Inference Images}
\begin{figure}[ht]
	\centering 
	\includegraphics[width=0.5\textwidth]{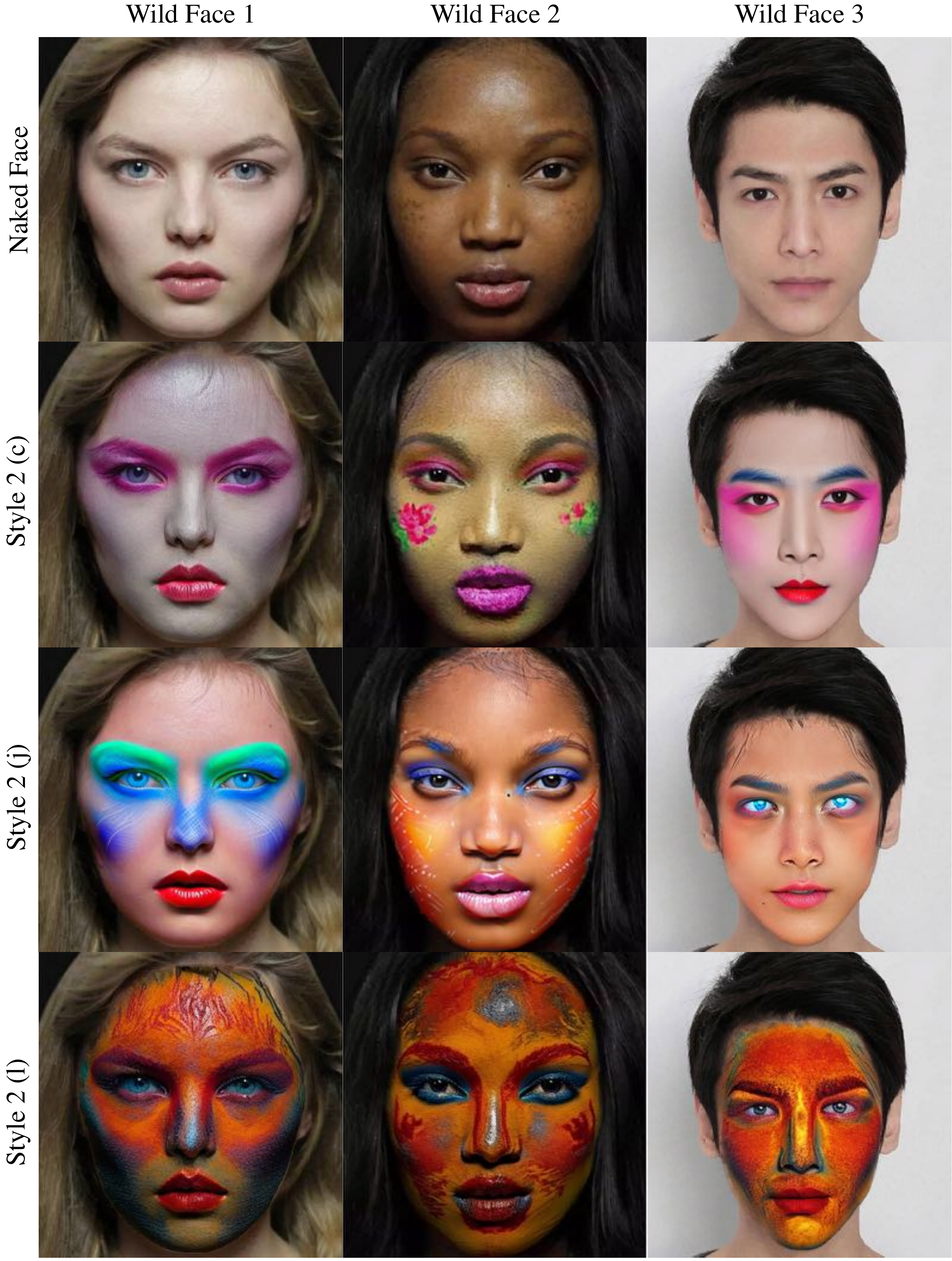}
	\caption{Application of $Gorgeous$ on wild inference images, demonstrating the versatility of our method in creating character makeup inspired by Style 2(c), 2(j), and 2(l) from Fig. \ref{fig:style2_bank}. The figure illustrates the successful adaptation of $Gorgeous$ to a variety of facial images outside the standard dataset, showcasing its potential for real-world applications.}
	\label{fig:wild}
\end{figure}

To further validate the adaptability and robustness of $Gorgeous$, we tested the method on wild inference images, which were collected randomly from online sources\footnote{\url{https://www.pinterest.com/}}\footnote{\url{https://k.sina.cn/}}\footnote{\url{https://www.facebook.com/}} and are not part of our standard dataset. This experiment is crucial for demonstrating how well $Gorgeous$ performs in real-world scenarios.
Figure \ref{fig:wild} showcases the capability of $Gorgeous$ to apply distinct and appropriate character makeup on these non-dataset facial images. The results highlight the method's effectiveness in handling a diverse array of facial features and expressions, adjusting the makeup application to suit individual characteristics without prior tuning. This ability makes $Gorgeous$ particularly valuable for practical applications where user-generated content varies widely in quality and style.
\begin{figure*}[ht]
	\centering 
	\includegraphics[width=1\textwidth]{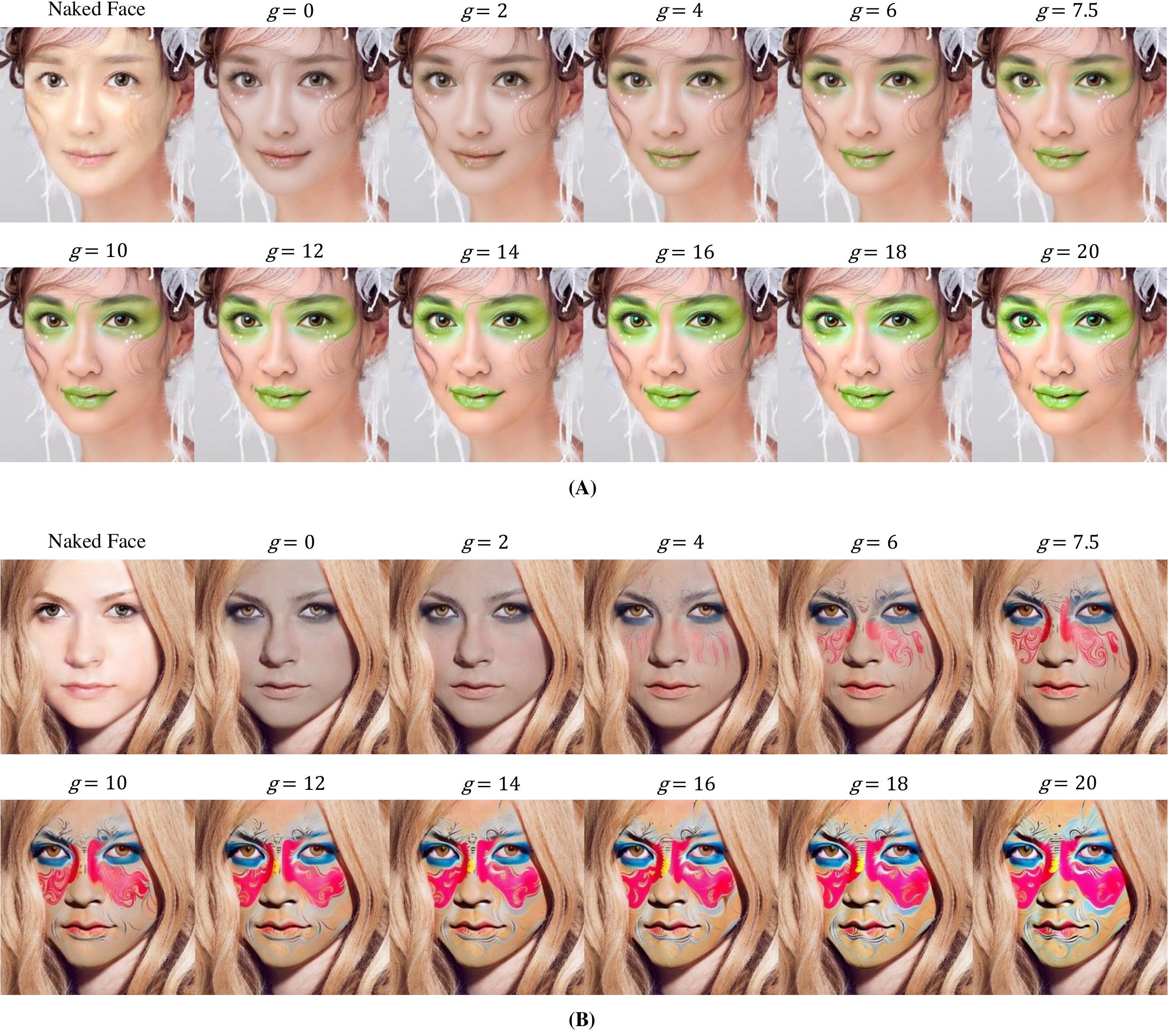}
	\caption{Impact of Varied Guidance Scales on Makeup Intensity. (A) depicts the effects of different guidance scales on a makeup style inspired by Style 1 (c), while (B) shows Style 1 (f) from Fig. \ref{fig:style1_bank}. Each panel presents a spectrum of intensities achieved by adjusting the guidance scale from 0 to 20, demonstrating the versatility of $Gorgeous$ in catering to personal preferences for makeup intensity.}
	\label{fig:guidance_scale}
\end{figure*}

\section{Further Analysis on Varied Implementation Details}
In this section, we delve deeper into the specific operational details of $Gorgeous$, particularly focusing on varied guidance scales and inference steps used during inference stage. 

\subsection{Adjustments in Guidance Scale and Inference Steps}
\begin{figure*}[ht]
	\centering 
	\includegraphics[width=1\textwidth]{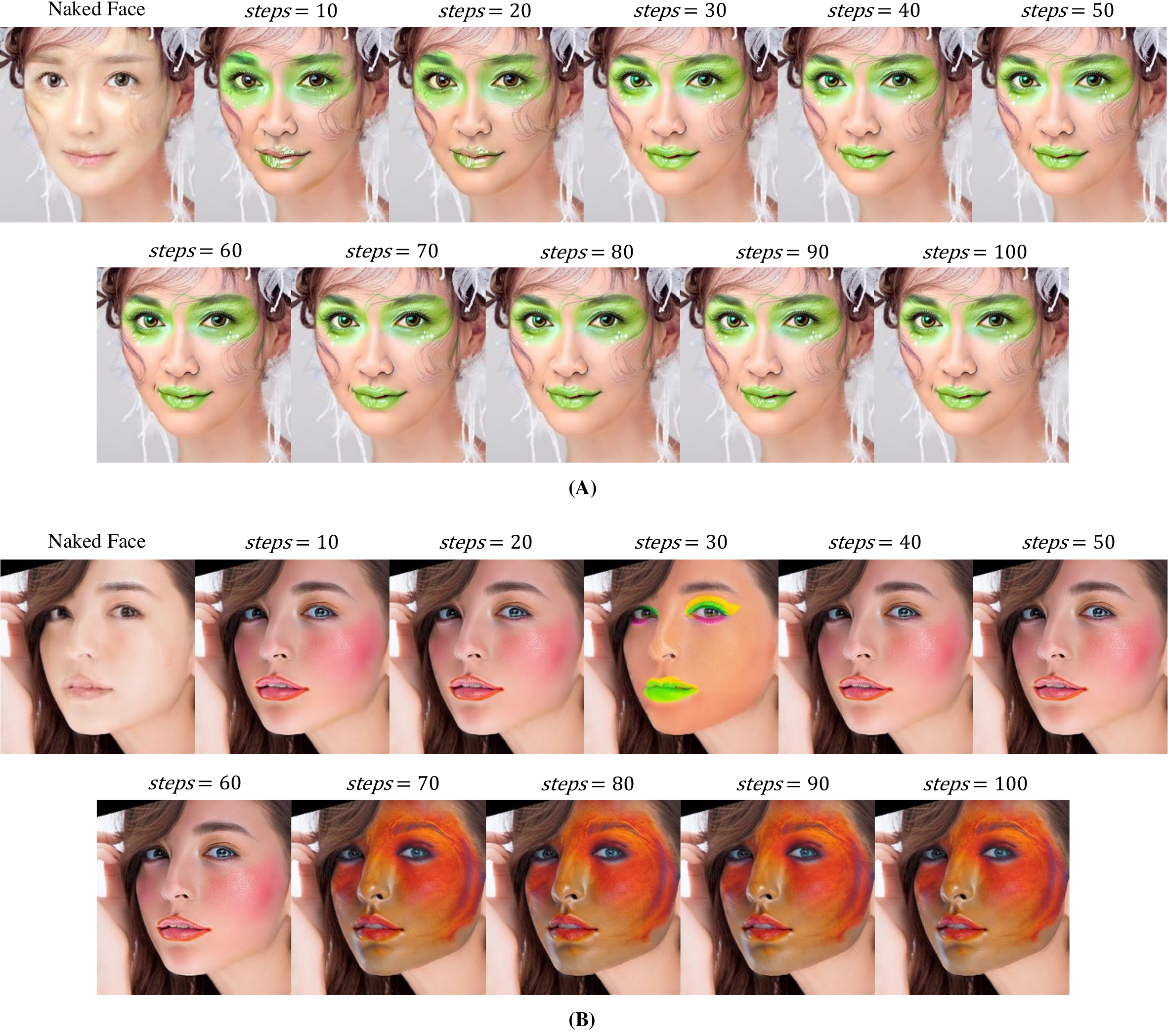}
	\caption{Influence of Inference Steps on Makeup Application. This figure illustrates the effect of varying the number of inference steps from 10 to 100 during the makeup generation process with $Gorgeous$. (A) reflects the makeup application inspired by Style 1 (c), while Panel B corresponds to Style 2 (1). Each panel reveals how the number of steps can alter the definition and vividness of the makeup, providing users with control over the gradual transformation from minimalist to more detailed makeup looks.}
	\label{fig:inference_steps}
\end{figure*}

As highlighted in the main text, the guidance scale $g$ is adjustable and ranges from 3 to 20, allowing for flexibility in the intensity and fidelity of the makeup application relative to the original inspiration (as depicted in Fig. \ref{fig:guidance_scale}). Additionally, the number of inference steps can vary between 30 to 100, depending on how pronounced or subtle the makeup needs to be (as shown in Fig. \ref{fig:inference_steps}). These parameters are crucial for tailoring the output to meet specific aesthetic goals, providing a versatile toolkit for makeup artists and designers to experiment with various visual outcomes.

\clearpage
\twocolumn
\section{Additional Ablation Study}
\begin{figure}[ht]
	\centering 
	\includegraphics[width=0.5\textwidth]{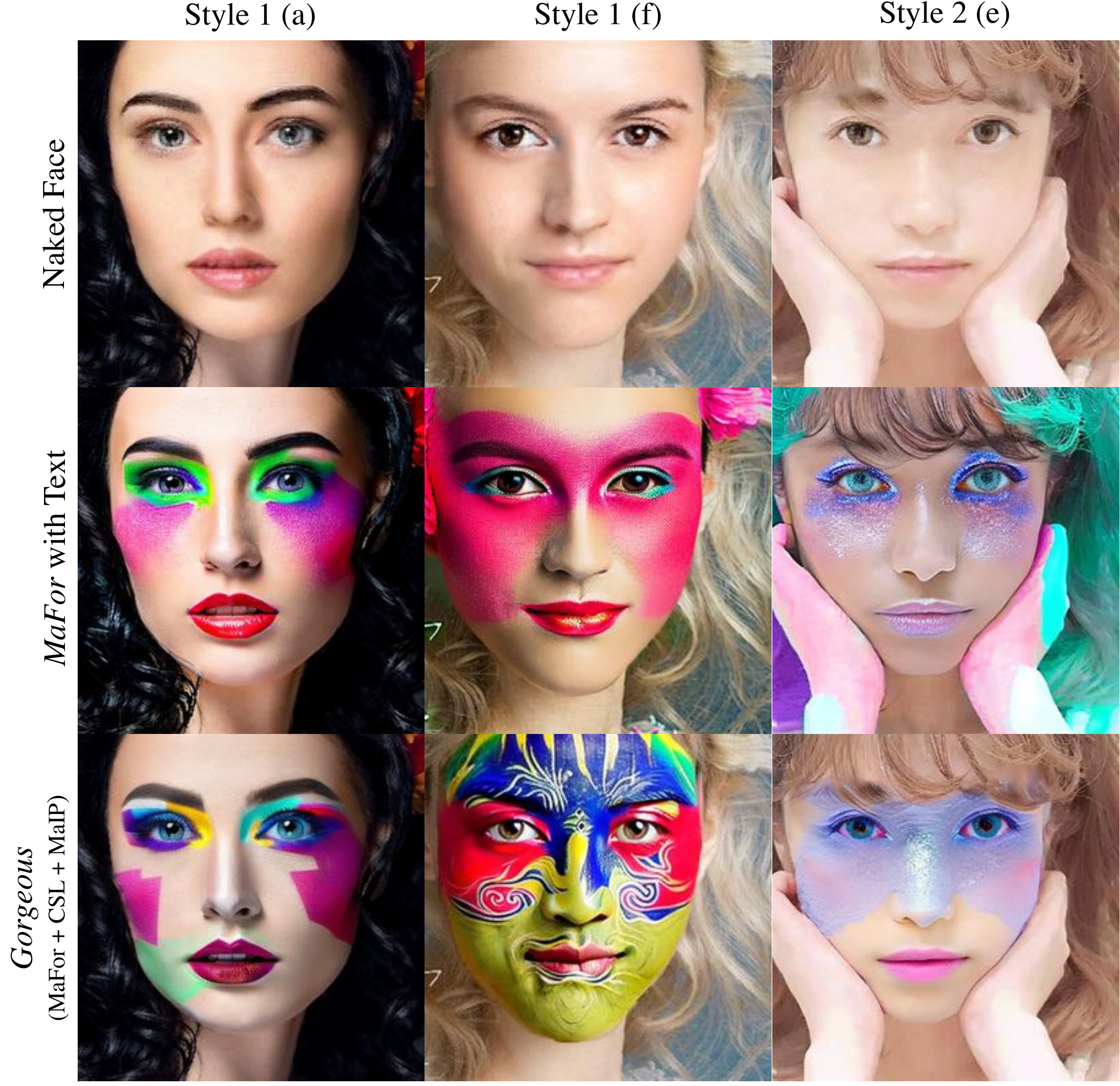}
	\caption{Ablation study showcasing the effectiveness of the CSL Module in $Gorgeous$. The visual outcomes for Style 1 (a), Style 1 (f), and Style 2 (e) demonstrate the nuanced interpretation of makeup styles facilitated by CSL. The figure contrasts the richly detailed and thematic makeup applications made possible with CSL against the more limited results achieved using only text prompts.}
	\label{fig:ablation}

\end{figure}

Figure \ref{fig:ablation} demonstrates the crucial role of the Character Settings Learning (CSL) Module in $Gorgeous$, illustrating its impact on the creation of character-specific facial makeup. The CSL Module enables $Gorgeous$ to go beyond the constraints of text-only prompts, as mentioned in the main text by utilizing themed reference images for generating makeup. This integration allows for a more accurate and thematic interpretation of makeup styles, such as specific art elements in Style 1 (a), Peking opera concepts in Style 1 (f), and natural textures like ice in Style 2 (e).

Without the CSL module, our system relies solely on text prompts, which significantly limits its ability to incorporate complex thematic concepts into the makeup designs. This often results in a generic or misaligned interpretation of the desired aesthetic, underscoring the module's importance in translating complex thematic elements from inspirational reference images into precise and meaningful makeup applications. The CSL module's capability to process and interpret diverse visual themes into practical makeup applications marks a substantial advancement in digital makeup technology, enhancing both the creativity and applicability of $Gorgeous$.

\section{Executive Summary}
In this supplementary material, we have expanded upon the initial findings presented in the main document, offering detailed insights into the implementation, performance, and adaptability of $Gorgeous$. The additional analyses, including extended quantitative and qualitative evaluations, demonstrate the robustness of our method across a diverse range of makeup applications.

The inclusion of wild inference images and varied implementation details further substantiates the versatility of $Gorgeous$, showcasing its ability to handle not just standard dataset images but also real-world, diverse facial images with high efficacy. 

Future efforts will focus on exploring advanced evaluation metrics specifically tailored for makeup application. These developments will aim to enhance the precision and user-specific adaptability of our system, ensuring that $Gorgeous$ remains at the forefront of digital makeup technology.

By providing these comprehensive analyses and detailed comparisons, we hope to contribute significantly to the ongoing advancements in digital makeup application, offering both theoretical insights and practical implications that could benefit researchers, developers, and end-users in related fields.


\end{document}